\documentclass[10pt]{article}
\usepackage[preprint]{tmlr}
\usepackage{hyperref}
\usepackage{url}
\usepackage{xurl}
\usepackage{microtype}
\usepackage{booktabs}

\usepackage{placeins}
\usepackage{etoolbox}
\AtBeginEnvironment{table}{\setlength{\abovecaptionskip}{0pt}\setlength{\belowcaptionskip}{6pt}}
\usepackage{amsmath,amssymb}
\usepackage{graphicx}
\usepackage{xcolor}
\hypersetup{hidelinks,pdfauthor={Junwei Yao, Yongxi Zhou, Zike Yang, Zhijun Wang, Jianan Liu, Yanxian Chen, Wenbo Ye},pdftitle={One Confidence Threshold, Unequal Coverage: Language Gaps in Typed Decision Interfaces and How Translating Inputs Reduces Them}}

\newcommand{\MlNRefused}{42}
\newcommand{\MlNRequests}{7,650}
\newcommand{\LocNRefused}{74}

\newcommand{\XnNRefused}{0}
\newcommand{\XnNRequests}{4,500}
\newcommand{\MlNLangs}{51}
\newcommand{\MlNPerLang}{150}

\newcommand{\MlAutoSeventyLo}{43}
\newcommand{\MlAutoSeventyHi}{73}

\newcommand{\MlAutoNinetyFiveLo}{15}
\newcommand{\MlAutoNinetyFiveHi}{53}

\newcommand{\MlAutoErrLo}{0}
\newcommand{\MlAutoErrHi}{8}

\newcommand{\MlHandoffHiLang}{jv}

\newcommand{\MlNAutoBelowEn}{8}

\newcommand{\MlAutoLoSw}{23}
\newcommand{\MlAutoHiSw}{37}

\newcommand{\MlAutoEnLo}{52}
\newcommand{\MlAutoEnHi}{68}
\newcommand{\MlNBothRightHandoffHi}{92}

\newcommand{\MlNEceDiffResolved}{2}

\newcommand{\MlEceDiffHiMax}{11}

\newcommand{\MlNAutoDiffResolved}{37}

\newcommand{\MlEceDiffSimQ}{3.48}
\newcommand{\MlAutoDiffSimQ}{3.18}

\newcommand{\MlOwnOverRateMean}{47}
\newcommand{\MlNOwnOverMajority}{17}
\newcommand{\MlEnOverRateMean}{69}
\newcommand{\MlOwnOverRateEn}{56}
\newcommand{\MlOwnOverRateLo}{22}
\newcommand{\MlOwnOverRateHi}{64}

\newcommand{\MlNRepSplits}{200}
\newcommand{\MlOwnOverOrInfeasibleMean}{47}
\newcommand{\MlNRepHalf}{75}
\newcommand{\MlNTune}{83}
\newcommand{\MlNTest}{67}

\newcommand{\MlEnThrErrHi}{11}
\newcommand{\MlEnThrErrHiLang}{sq}
\newcommand{\MlEnThrErrEn}{0}
\newcommand{\MlNEnThrOverTarget}{12}
\newcommand{\MlNEnThrOverDouble}{1}

\newcommand{\MlNOwnOverTarget}{17}

\newcommand{\MlNOwnHold}{34}
\newcommand{\MlOwnCovHoldLo}{31}
\newcommand{\MlOwnCovHoldHi}{79}

\newcommand{\MlLttMinUcbEn}{5.3}
\newcommand{\MlLttMinUcbLo}{4.2}
\newcommand{\MlLttMinUcbHi}{14.2}
\newcommand{\MlLttMinUcbLoLang}{pt}
\newcommand{\MlLttMinUcbHiLang}{sq}

\newcommand{\MlNBudgetTFOverDouble}{5}

\newcommand{\MlNBudgetFOverDouble}{1}
\newcommand{\XnNLangs}{15}
\newcommand{\XnNPerLang}{300}

\newcommand{\XnAutoErrHi}{29}

\newcommand{\XnAutoErrHiLang}{ur}

\newcommand{\XnNEceDiffLangs}{14}
\newcommand{\XnNEceDiffResolved}{11}

\newcommand{\XnNEceDiffResolvedSim}{10}

\newcommand{\XnEceDiffSimQ}{2.88}
\newcommand{\XnAutoDiffSimQ}{2.82}

\newcommand{\XnNTune}{156}
\newcommand{\XnNTest}{144}

\newcommand{\XnNEnThrOverTarget}{15}

\newcommand{\XnNOwnOverTarget}{3}
\newcommand{\XnNOwnFeasible}{3}

\newcommand{\XnLttMinUcbLo}{8.1}
\newcommand{\XnLttMinUcbHi}{33.9}

\newcommand{\LyMlAccLo}{6}
\newcommand{\LyMlAccHi}{55}
\newcommand{\LyMlAccEn}{55}

\newcommand{\LyMlAutoLo}{1}
\newcommand{\LyMlAutoHi}{37}

\newcommand{\LyXnEceHi}{32}

\newcommand{\LyXnAutoLo}{71}
\newcommand{\LyXnAutoHi}{83}

\newcommand{\LyXnAutoErrHi}{34}
\newcommand{\LyXnAutoErrEn}{9}

\newcommand{\LocAutoLo}{16}
\newcommand{\LocAutoHi}{59}

\newcommand{\LocAutoSw}{35}

\newcommand{\LocAutoEn}{56}

\newcommand{\LocDAccMn}{+1}

\newcommand{\LocMeanDAuto}{\ensuremath{-}7.6}

\newcommand{\LocNAutoUp}{16}
\newcommand{\LocNAutoDown}{28}

\newcommand{\LocGapEnSwEn}{25}
\newcommand{\LocGapEnSwLoc}{21}
\newcommand{\LocSpearmanAuto}{0.19}
\newcommand{\LocErrLo}{0}
\newcommand{\LocErrHi}{6}
\newcommand{\LocSpreadEn}{31}
\newcommand{\LocSpreadLoc}{40}

\newcommand{\LocMaxLossLang}{mn}

\newcommand{\LocMeanDAutoBottom}{\ensuremath{-}0.1}
\newcommand{\LocMeanDAutoRest}{\ensuremath{-}9.5}
\newcommand{\LocDAccAbsMax}{5}

\newcommand{\LocCorrDAccDAuto}{0.20}

\newcommand{\MlXBosunRho}{0.76}

\newcommand{\MlXBosunBottomOverlap}{6}

\newcommand{\MlXDeciderRho}{0.54}
\newcommand{\MlXDeciderEn}{45}

\newcommand{\MlXDeciderAccEn}{85}
\newcommand{\MlXDeciderBottomOverlap}{5}

\newcommand{\MlXLummaRho}{0.47}
\newcommand{\MlXLummaEn}{13}

\newcommand{\MlXLummaAccEn}{48}
\newcommand{\MlXLummaBottomOverlap}{5}
\newcommand{\MlXLummaBottomSize}{12}

\newcommand{\MlXSvcNBelowEn}{8}
\newcommand{\MlXBosunNBelowEn}{25}
\newcommand{\MlXDeciderNBelowEn}{36}
\newcommand{\MlXLummaNBelowEn}{38}

\newcommand{\MlXXBosunMax}{3}

\newcommand{\MlXXDeciderRho}{0.77}
\newcommand{\MlXXDeciderEn}{60}

\newcommand{\MlXXSvcEn}{68}

\newcommand{\MlLttNeedOne}{45}
\newcommand{\MlLttNeedGrid}{122}
\newcommand{\MlLttGridN}{50}

\newcommand{\pxAlpha}{90}
\newcommand{\pxMiss}{10}
\newcommand{\pxNReps}{200}

\newcommand{\pxMlNCalPooled}{3,825}
\newcommand{\pxMlNCal}{75}

\newcommand{\pxMlSingMondLo}{5}
\newcommand{\pxMlSingMondHi}{61}
\newcommand{\pxMlSingMondLoLang}{jv}
\newcommand{\pxMlSingMondHiLang}{ja}
\newcommand{\pxMlSingMondEn}{56}

\newcommand{\pxMlSingErrMondHi}{5}

\newcommand{\pxMlCovPooledRepLo}{84}

\newcommand{\pxMlCovPooledRepLoLang}{km}

\newcommand{\pxMlCovMondRepLo}{91}
\newcommand{\pxMlCovMondRepHi}{96}

\newcommand{\pxMlDeferCorrectMondEn}{28}
\newcommand{\pxMlDeferCorrectMondSw}{61}

\newcommand{\pxMlGateDeferCorrectEn}{23}
\newcommand{\pxMlGateDeferCorrectSw}{39}
\newcommand{\pxMlSingPooledSpread}{30}
\newcommand{\pxMlSingMondSpread}{51}
\newcommand{\pxMlGateSpread}{36}

\newcommand{\pxMlCovPooledRepMean}{91}

\newcommand{\pxMlSizeMondHi}{21.5}
\newcommand{\pxMlSizeMondEn}{3.0}
\newcommand{\pxMlSizeMondParHi}{20.5}

\newcommand{\pxMlSizeMondParEn}{2.5}
\newcommand{\pxMlNQOwnMax}{42}

\newcommand{\pxMlPKeepMin}{0.01}
\newcommand{\pxMlNCovPooledBelow}{16}

\newcommand{\pxMlCovMondRepMin}{91}
\newcommand{\pxMlNLangs}{51}
\newcommand{\pxMlBotGateSize}{10}
\newcommand{\pxMlBotCovPooledSize}{10}
\newcommand{\pxMlBotCovPooledOverlap}{5}

\newcommand{\pxMlBotSingMondSize}{10}

\newcommand{\pxXnSingErrMondHi}{24}

\newcommand{\pxXnSingErrMondHiLang}{ur}

\newcommand{\pxXnSingPooledSpread}{25}
\newcommand{\pxXnSingMondSpread}{47}
\newcommand{\pxXnGateSpread}{21}

\newcommand{\pxXnNCovPooledBelow}{7}

\newcommand{\pxXnCovMondRepMin}{90}
\newcommand{\pxXnNLangs}{15}

\newcommand{\pxLocSingMondSpread}{43}

\newcommand{\pxLocBotSingMondSize}{10}
\newcommand{\pxLocBotSingMondOverlap}{6}

\newcommand{\pbtNLangs}{50}
\newcommand{\pbtNItems}{60}
\newcommand{\pbtModel}{gpt-5.4-mini}

\newcommand{\pbtJudgeMean}{4.71}
\newcommand{\pbtShareLeTwo}{5.8}
\newcommand{\pbtNLangsLowItem}{42}
\newcommand{\pbtShareLowTopTen}{54}
\newcommand{\pbtNInstrLow}{6}

\newcommand{\pbtRhoJudgeAuto}{\ensuremath{-}0.14}
\newcommand{\pbtPJudgeAuto}{0.35}

\newcommand{\pbtRhoLexAuto}{\ensuremath{-}0.26}
\newcommand{\pbtPLexAuto}{0.070}
\newcommand{\pbtRhoLexAutoLabels}{\ensuremath{-}0.26}
\newcommand{\pbtPLexAutoLabels}{0.072}
\newcommand{\pbtNTests}{7}
\newcommand{\pbtPLexAutoLabelsBonf}{0.51}

\newcommand{\pbtDriftRankMn}{14}

\newcommand{\swfixN}{150}
\newcommand{\swfixNEntries}{80}

\newcommand{\swfixNFromReviews}{24}
\newcommand{\swfixNFromAzure}{56}

\newcommand{\swfixGapEn}{25.3}

\newcommand{\swfixAutoOrig}{34.7}
\newcommand{\swfixAccOrig}{70.7}
\newcommand{\swfixEceOrig}{7}
\newcommand{\swfixGapOrig}{21.3}

\newcommand{\swfixAutoFix}{36.0}
\newcommand{\swfixAccFix}{72.7}
\newcommand{\swfixEceFix}{8}
\newcommand{\swfixGapFix}{20.0}

\newcommand{\swfixDAutoOrig}{+1.3}
\newcommand{\swfixDAutoOrigLo}{\ensuremath{-}5.3}
\newcommand{\swfixDAutoOrigHi}{+7.3}

\newcommand{\swfixDAccOrigLo}{\ensuremath{-}2.0}
\newcommand{\swfixDAccOrigHi}{+6.0}

\newcommand{\swfixDEceOrigLo}{\ensuremath{-}4}
\newcommand{\swfixDEceOrigHi}{+5}

\newcommand{\phSeventyMeanGap}{6}
\newcommand{\phSeventyMeanGapLo}{3}
\newcommand{\phSeventyMeanGapHi}{10}
\newcommand{\phSeventySwGap}{17}
\newcommand{\phSeventySwLo}{4}
\newcommand{\phSeventySwHi}{30}

\newcommand{\phSeventyNBelowSim}{9}

\newcommand{\phEightyMeanGap}{9}
\newcommand{\phEightyMeanGapLo}{5}
\newcommand{\phEightyMeanGapHi}{13}
\newcommand{\phEightySwGap}{25}
\newcommand{\phEightySwLo}{11}
\newcommand{\phEightySwHi}{38}

\newcommand{\phEightyNBelowSim}{15}

\newcommand{\phNinetyMeanGap}{12}
\newcommand{\phNinetyMeanGapLo}{8}
\newcommand{\phNinetyMeanGapHi}{17}
\newcommand{\phNinetySwGap}{31}
\newcommand{\phNinetySwLo}{17}
\newcommand{\phNinetySwHi}{44}

\newcommand{\phNinetyNBelowSim}{23}

\newcommand{\phNinetyFiveMeanGap}{15}
\newcommand{\phNinetyFiveMeanGapLo}{10}
\newcommand{\phNinetyFiveMeanGapHi}{20}
\newcommand{\phNinetyFiveSwGap}{32}
\newcommand{\phNinetyFiveSwLo}{19}
\newcommand{\phNinetyFiveSwHi}{45}

\newcommand{\phNinetyFiveNBelowSim}{35}

\newcommand{\phSimB}{20,000}

\newcommand{\phJoshiRho}{\ensuremath{-}0.57}
\newcommand{\phJoshiRhoLo}{\ensuremath{-}0.74}
\newcommand{\phJoshiRhoHi}{\ensuremath{-}0.22}
\newcommand{\phJoshiN}{50}
\newcommand{\phJoshiGapB}{17}
\newcommand{\phJoshiNB}{12}
\newcommand{\phJoshiGapC}{22}
\newcommand{\phJoshiNC}{3}
\newcommand{\phJoshiGapD}{12}
\newcommand{\phJoshiND}{15}
\newcommand{\phJoshiGapE}{7}
\newcommand{\phJoshiNE}{13}
\newcommand{\phJoshiGapF}{8}
\newcommand{\phJoshiNF}{7}

\newcommand{\phMlFsNCertified}{0}
\newcommand{\phMlFsNLangs}{51}
\newcommand{\phMlFsNTopMax}{26}
\newcommand{\phXnFsNCertified}{0}
\newcommand{\phXnFsNLangs}{15}

\newcommand{\phXnRhoEceCov}{\ensuremath{-}0.41}
\newcommand{\phXnRhoEceCovLo}{\ensuremath{-}0.84}
\newcommand{\phXnRhoEceCovHi}{0.37}
\newcommand{\phXnNLangs}{15}

\newcommand{\phMlEnFitThr}{0.74}
\newcommand{\phXnEnFitThr}{0.98}
\newcommand{\phNBoot}{2,000}

\newcommand{\pcoMlAutoEnCo}{60}

\newcommand{\pcoMlAutoSwCo}{29}
\newcommand{\pcoMlAutoJvCo}{24}

\newcommand{\pcoMlNAutoBelowHalfCo}{26}

\newcommand{\pcoMlNEceDiffResolvedCo}{2}
\newcommand{\pcoMlNEceDiffResolvedSimCo}{0}

\newcommand{\pcoMlNOwnHoldCo}{34}

\newcommand{\pcoMlLttNFeasibleCo}{0}

\newcommand{\pcoMlNLangsCo}{51}

\newcommand{\pcoMlSeventyNBelowSimCo}{9}
\newcommand{\pcoMlSeventyNAboveSimCo}{0}

\newcommand{\pcoMlEightyNBelowSimCo}{15}
\newcommand{\pcoMlEightyNAboveSimCo}{0}

\newcommand{\pcoMlNinetyNBelowSimCo}{23}
\newcommand{\pcoMlNinetyNAboveSimCo}{0}
\newcommand{\pcoMlNinetyMeanGapCo}{12}
\newcommand{\pcoMlNinetyMeanGapLoCo}{8}
\newcommand{\pcoMlNinetyMeanGapHiCo}{17}

\newcommand{\pcoMlNinetyFiveNBelowSimCo}{35}
\newcommand{\pcoMlNinetyFiveNAboveSimCo}{0}

\newcommand{\pcoMlAutoEnMq}{59}

\newcommand{\pcoMlAutoSwMq}{31}
\newcommand{\pcoMlAutoJvMq}{23}

\newcommand{\pcoMlNAutoBelowHalfMq}{30}

\newcommand{\pcoMlNEceDiffResolvedMq}{4}
\newcommand{\pcoMlNEceDiffResolvedSimMq}{0}

\newcommand{\pcoMlNOwnHoldMq}{34}

\newcommand{\pcoMlLttNFeasibleMq}{0}

\newcommand{\pcoMlSeventyNBelowSimMq}{12}
\newcommand{\pcoMlSeventyNAboveSimMq}{0}

\newcommand{\pcoMlEightyNBelowSimMq}{14}
\newcommand{\pcoMlEightyNAboveSimMq}{0}

\newcommand{\pcoMlNinetyNBelowSimMq}{19}
\newcommand{\pcoMlNinetyNAboveSimMq}{0}
\newcommand{\pcoMlNinetyMeanGapMq}{11}
\newcommand{\pcoMlNinetyMeanGapLoMq}{7}
\newcommand{\pcoMlNinetyMeanGapHiMq}{16}

\newcommand{\pcoMlNinetyFiveNBelowSimMq}{30}
\newcommand{\pcoMlNinetyFiveNAboveSimMq}{0}

\newcommand{\pcoMlCmpTvMean}{0.019}

\newcommand{\pcoMlCmpRhoCov}{0.96}
\newcommand{\pcoMlCmpRhoGap}{0.96}
\newcommand{\pcoMlCmpTopAgree}{98.0}
\newcommand{\pcoMlCmpGateAgree}{98.1}

\newcommand{\pcoMlCmpDCovMean}{\ensuremath{-}0.1}
\newcommand{\pcoMlCmpDCovMeanLo}{\ensuremath{-}0.4}
\newcommand{\pcoMlCmpDCovMeanHi}{0.2}

\newcommand{\pcoMlCmpDCovMin}{\ensuremath{-}3.3}
\newcommand{\pcoMlCmpDCovMax}{2.7}

\newcommand{\pcoMlCmpNDCovResolvedSim}{0}
\newcommand{\pcoMlCmpNPairs}{7,650}
\newcommand{\pcoMlCmpNPairsAnswered}{7,601}
\newcommand{\pcoXnAutoEnCo}{68}

\newcommand{\pcoXnAutoUrCo}{47}

\newcommand{\pcoXnNEceDiffAboveSimCo}{10}

\newcommand{\pcoXnNDiffLangsCo}{14}

\newcommand{\pcoXnAutoEnMq}{68}

\newcommand{\pcoXnAutoUrMq}{47}

\newcommand{\pcoXnNEceDiffAboveSimMq}{9}

\newcommand{\pcoXnCmpTvMean}{0.011}

\newcommand{\pcoXnCmpRhoCov}{0.98}
\newcommand{\pcoXnCmpRhoGap}{0.98}
\newcommand{\pcoXnCmpTopAgree}{98.9}
\newcommand{\pcoXnCmpGateAgree}{98.0}

\newcommand{\pcoXnCmpDCovMean}{0.1}
\newcommand{\pcoXnCmpDCovMeanLo}{\ensuremath{-}0.3}
\newcommand{\pcoXnCmpDCovMeanHi}{0.5}

\newcommand{\pcoXnCmpNDCovResolvedSim}{0}
\newcommand{\pcoXnCmpNPairs}{4,500}
\newcommand{\pcoXnCmpNPairsAnswered}{4,500}

\newcommand{\pcoRepN}{506}
\newcommand{\pcoRepTvMean}{0.020}

\newcommand{\pcoRepTopAgree}{97.4}
\newcommand{\pcoRepGateAgree}{98.6}

\newcommand{\pcoRepCoTvMean}{0.021}

\newcommand{\pcoRepCoTopAgree}{96.2}
\newcommand{\pcoRepCoGateAgree}{98.6}
\newcommand{\pcoLocEnN}{150}
\newcommand{\pcoLocEnSame}{150}
\newcommand{\pcoRerunN}{7,589}
\newcommand{\pcoRerunTopAgree}{94.5}
\newcommand{\pcoRerunGateAgree}{93.8}

\newcommand{\exGold}{\texttt{datetime\_query}}
\newcommand{\exUttEn}{hey siri what{'}s the time}
\newcommand{\exPEn}{0.98}
\newcommand{\exUttSw}{hey siri saa ngapi}
\newcommand{\exPSw}{0.52}
\newcommand{\exUttJv}{heh siri jam pira}
\newcommand{\exPJv}{0.30}
\newcommand{\rkSvcEnRank}{1}

\newcommand{\rkSvcEnOverMedian}{11}

\newcommand{\rkDeciderEnRank}{1}

\newcommand{\rkDeciderEnOverMedian}{25}

\newcommand{\rkBosunEnRank}{2}

\newcommand{\rkBosunEnOverMedian}{16}
\newcommand{\rkBosunAboveEn}{zh-CN}
\newcommand{\rkLummaEnRank}{1}
\newcommand{\rkLummaEnTies}{1}
\newcommand{\rkLummaEnOverMedian}{11}

\newcommand{\rcvMlGateSpread}{36}

\newcommand{\rcvMlFiveEn}{80}

\newcommand{\rcvMlFiveMeanGap}{14}
\newcommand{\rcvMlFiveMeanGapLo}{6}
\newcommand{\rcvMlFiveMeanGapHi}{24}

\newcommand{\rcvMlFiveThrLo}{0.44}
\newcommand{\rcvMlFiveThrHi}{0.94}
\newcommand{\rcvMlFiveThrEn}{0.52}

\newcommand{\rcvMlHalfEn}{0}

\newcommand{\rcvMlHalfHi}{9}
\newcommand{\rcvMlHalfHiLang}{sq}

\newcommand{\rcvMlHalfMeanGap}{2}
\newcommand{\rcvMlHalfMeanGapLo}{1}
\newcommand{\rcvMlHalfMeanGapHi}{4}

\newcommand{\rcvXnTenEn}{60}

\newcommand{\rcvXnTenNReach}{2}

\newcommand{\rcvXnHalfEn}{8}

\newcommand{\rcvXnHalfHi}{29}
\newcommand{\rcvXnHalfHiLang}{ur}

\newcommand{\rcvXnHalfNAboveSim}{13}

\newcommand{\rcvMlDeciderFiveEn}{75}
\newcommand{\rcvMlDeciderFiveLo}{0}
\newcommand{\rcvMlDeciderFiveHi}{75}

\newcommand{\rcvMlDeciderFiveNBelowSim}{21}
\newcommand{\rcvMlBosunFiveEn}{45}
\newcommand{\rcvMlBosunFiveLo}{0}
\newcommand{\rcvMlBosunFiveHi}{50}

\newcommand{\rcvMlBosunFiveNBelowSim}{0}

\newcommand{\rcvXBosunErrEn}{7}
\newcommand{\rcvXBosunErrOther}{14}

\newcommand{\rcvXLummaErrEn}{5}
\newcommand{\rcvXLummaErrOther}{22}

\newcommand{\rcvMlOwnNTest}{67}
\newcommand{\rcvMlOwnWorstLang}{nb}

\newcommand{\rcvXnOwnNTest}{144}
\newcommand{\rcvXnOwnWorstLang}{el}

\newcommand{\rcvApsMlSingPooledSpread}{30}

\newcommand{\rcvApsMlNCovPooledBelow}{24}
\newcommand{\rcvApsMlCovPooledMin}{85}

\newcommand{\rcvApsMlSingMondSpread}{35}
\newcommand{\rcvApsMlSingMondLo}{21}
\newcommand{\rcvApsMlSingMondLoLang}{jv}

\newcommand{\rcvApsXnSingPooledSpread}{27}

\newcommand{\rcvApsXnNCovPooledBelow}{9}

\newcommand{\rcvApsXnSingMondSpread}{41}

\newcommand{\rcvNBoot}{10,000}

\newcommand{\pphRhoPara}{0.88}

\newcommand{\pphMeanDPara}{+2.5}

\newcommand{\pphGateAgreePara}{85}

\newcommand{\pphMeanDAccPara}{+0.4}
\newcommand{\pphErrHiPara}{10}

\newcommand{\pphGateAgreeLoc}{81}

\newcommand{\pphRhoDiff}{0.69}
\newcommand{\pphRhoDiffLo}{0.45}
\newcommand{\pphRhoDiffHi}{0.76}

\newcommand{\pphNRefused}{9}

\newcommand{\cfNItems}{1{,}000}
\newcommand{\cfNReq}{51{,}000}
\newcommand{\cfB}{10{,}000}

\newcommand{\cfLunaDelta}{11.2}
\newcommand{\cfLunaDeltaLo}{9.0}
\newcommand{\cfLunaDeltaHi}{13.3}
\newcommand{\cfLunaPHolm}{<0.001}

\newcommand{\cfLunaAnswered}{99.7}
\newcommand{\cfLunaNCovBelow}{45}
\newcommand{\cfLunaNCovAbove}{0}
\newcommand{\cfLunaNEceWorse}{0}
\newcommand{\cfLunaNEceBetter}{0}
\newcommand{\cfLunaNHgcBelow}{44}

\newcommand{\cfLunaStd}{8.5}
\newcommand{\cfLunaStdLo}{6.6}
\newcommand{\cfLunaStdHi}{10.4}
\newcommand{\cfLunaNPdPos}{50}
\newcommand{\cfLunaPdMin}{2}
\newcommand{\cfLunaPdMinLang}{ja}
\newcommand{\cfLunaPdMax}{37}
\newcommand{\cfLunaPdMaxLang}{jv}
\newcommand{\cfLunaCovEn}{59}
\newcommand{\cfLunaCovSw}{36}
\newcommand{\cfLunaCovJv}{27}
\newcommand{\cfLunaCovMin}{27}
\newcommand{\cfLunaCovMinLang}{jv}

\newcommand{\cfLunaCovGapMax}{32}
\newcommand{\cfLunaAccEn}{78}
\newcommand{\cfLunaAccMin}{60}

\newcommand{\cfLunaSelErrEn}{5.1}
\newcommand{\cfLunaSelErrMax}{8.1}
\newcommand{\cfLunaSelErrMaxLang}{jv}
\newcommand{\cfLunaSelErrUcbMax}{12.0}
\newcommand{\cfLunaSelErrUcbMaxLang}{jv}
\newcommand{\cfLunaSelErrUcbMaxErr}{22}
\newcommand{\cfLunaSelErrUcbMaxHandled}{272}
\newcommand{\cfLunaEceEn}{4.0}
\newcommand{\cfLunaEceMin}{2.2}
\newcommand{\cfLunaEceMax}{5.7}
\newcommand{\cfLunaEceGapMaxAbs}{1.8}
\newcommand{\cfLunaEceHalfMed}{3.6}
\newcommand{\cfLunaHgcEn}{72}
\newcommand{\cfLunaHgcSw}{51}
\newcommand{\cfLunaCdEn}{22}
\newcommand{\cfLunaCdSw}{32}

\newcommand{\cfLunaRho}{0.84}
\newcommand{\cfMsDelta}{11.7}
\newcommand{\cfMsDeltaLo}{10.2}
\newcommand{\cfMsDeltaHi}{13.2}
\newcommand{\cfMsPHolm}{<0.001}

\newcommand{\cfMsNCovBelow}{44}
\newcommand{\cfMsNCovAbove}{0}
\newcommand{\cfMsNEceWorse}{1}
\newcommand{\cfMsNEceBetter}{0}
\newcommand{\cfMsNHgcBelow}{34}

\newcommand{\cfMsStd}{9.0}
\newcommand{\cfMsStdLo}{7.3}
\newcommand{\cfMsStdHi}{10.7}
\newcommand{\cfMsNPdPos}{50}
\newcommand{\cfMsPdMin}{2}
\newcommand{\cfMsPdMinLang}{de}
\newcommand{\cfMsPdMax}{39}
\newcommand{\cfMsPdMaxLang}{ta}
\newcommand{\cfMsCovEn}{75}
\newcommand{\cfMsCovSw}{41}
\newcommand{\cfMsCovJv}{47}
\newcommand{\cfMsCovMin}{33}
\newcommand{\cfMsCovMinLang}{ta}

\newcommand{\cfMsAccEn}{88}
\newcommand{\cfMsAccMin}{58}

\newcommand{\cfMsSelErrEn}{3.6}
\newcommand{\cfMsSelErrMax}{7.0}
\newcommand{\cfMsSelErrMaxLang}{sq}

\newcommand{\cfMsEceEn}{3.6}
\newcommand{\cfMsEceMin}{2.4}
\newcommand{\cfMsEceMax}{8.8}

\newcommand{\cfMsHgcEn}{82}
\newcommand{\cfMsHgcSw}{54}
\newcommand{\cfMsCdEn}{16}
\newcommand{\cfMsCdSw}{33}

\newcommand{\cfPxDelta}{8.6}
\newcommand{\cfPxDeltaLo}{7.2}
\newcommand{\cfPxDeltaHi}{9.9}
\newcommand{\cfPxPHolm}{<0.001}

\newcommand{\cfPxNCovBelow}{47}
\newcommand{\cfPxNCovAbove}{0}
\newcommand{\cfPxNEceWorse}{6}
\newcommand{\cfPxNEceBetter}{0}
\newcommand{\cfPxNHgcBelow}{38}

\newcommand{\cfPxStd}{6.9}
\newcommand{\cfPxStdLo}{5.1}
\newcommand{\cfPxStdHi}{8.6}
\newcommand{\cfPxNPdPos}{50}
\newcommand{\cfPxPdMin}{2}
\newcommand{\cfPxPdMinLang}{ru}
\newcommand{\cfPxPdMax}{23}
\newcommand{\cfPxPdMaxLang}{am}
\newcommand{\cfPxCovEn}{78}
\newcommand{\cfPxCovSw}{48}
\newcommand{\cfPxCovJv}{55}
\newcommand{\cfPxCovMin}{48}
\newcommand{\cfPxCovMinLang}{cy}

\newcommand{\cfPxAccEn}{88}
\newcommand{\cfPxAccMin}{69}

\newcommand{\cfPxSelErrEn}{3.6}
\newcommand{\cfPxSelErrMax}{8.2}
\newcommand{\cfPxSelErrMaxLang}{sq}

\newcommand{\cfPxEceEn}{3.8}
\newcommand{\cfPxEceMin}{3.5}
\newcommand{\cfPxEceMax}{9.4}

\newcommand{\cfPxHgcEn}{85}
\newcommand{\cfPxHgcSw}{65}
\newcommand{\cfPxCdEn}{13}
\newcommand{\cfPxCdSw}{24}

\newcommand{\ttMTModel}{gpt-5.4-mini}
\newcommand{\ttNUtt}{50{,}000}
\newcommand{\ttDelta}{0.8}
\newcommand{\ttDeltaLo}{\ensuremath{-}1.1}
\newcommand{\ttDeltaHi}{2.6}
\newcommand{\ttDeltaOrig}{11.2}
\newcommand{\ttDeltaOrigLo}{9.0}
\newcommand{\ttDeltaOrigHi}{13.3}

\newcommand{\ttAnswered}{99.6}
\newcommand{\ttCovEn}{59}
\newcommand{\ttCovSw}{58}
\newcommand{\ttCovJv}{57}
\newcommand{\ttCovSwOrig}{36}
\newcommand{\ttCovJvOrig}{27}
\newcommand{\ttCovMin}{45}
\newcommand{\ttCovMinLang}{km}

\newcommand{\ttAccEn}{78}
\newcommand{\ttAccMin}{64}

\newcommand{\ttAccMean}{76}
\newcommand{\ttAccMeanOrig}{74}
\newcommand{\ttAccMinOrig}{60}

\newcommand{\ttSelErrMax}{7.8}
\newcommand{\ttSelErrMaxLang}{am}
\newcommand{\ttNCovBelow}{3}

\newcommand{\ttNCovBelowOrig}{45}

\newcommand{\ttCovMinOrig}{27}
\newcommand{\ttCovMinOrigLang}{jv}
\newcommand{\ttCovGapMin}{14}
\newcommand{\ttCovGapMinOrig}{32}
\newcommand{\ttSelErrMaxOrig}{8.1}
\newcommand{\ttSelErrMaxOrigLang}{jv}
\newcommand{\ttSelErrSw}{7.1}
\newcommand{\ttSelErrSwOrig}{5.8}

\newcommand{\ttIntDelta}{0.5}

\newcommand{\ttIntDeltaOrig}{11.0}

\newcommand{\ttIntDiffDelta}{\ensuremath{-}10.5}
\newcommand{\ttIntDiffDeltaLo}{\ensuremath{-}11.7}
\newcommand{\ttIntDiffDeltaHi}{\ensuremath{-}9.4}
\newcommand{\ttMsDelta}{1.0}
\newcommand{\ttMsDeltaLo}{\ensuremath{-}0.1}
\newcommand{\ttMsDeltaHi}{2.1}
\newcommand{\ttMsDeltaOrig}{11.7}
\newcommand{\ttMsDeltaOrigLo}{10.2}
\newcommand{\ttMsDeltaOrigHi}{13.2}

\newcommand{\ttMsCovEn}{75}
\newcommand{\ttMsCovSw}{72}
\newcommand{\ttMsCovJv}{73}
\newcommand{\ttMsCovSwOrig}{41}
\newcommand{\ttMsCovJvOrig}{47}
\newcommand{\ttMsCovMin}{57}
\newcommand{\ttMsCovMinLang}{km}

\newcommand{\ttMsAccEn}{88}
\newcommand{\ttMsAccMin}{74}

\newcommand{\ttMsAccMean}{85}
\newcommand{\ttMsAccMeanOrig}{80}
\newcommand{\ttMsAccMinOrig}{58}

\newcommand{\ttMsSelErrMax}{6.4}
\newcommand{\ttMsSelErrMaxLang}{th}
\newcommand{\ttMsNCovBelow}{8}

\newcommand{\ttMsNCovBelowOrig}{44}

\newcommand{\ttMsCovMinOrig}{33}
\newcommand{\ttMsCovMinOrigLang}{ta}
\newcommand{\ttMsCovGapMin}{17}
\newcommand{\ttMsCovGapMinOrig}{42}
\newcommand{\ttMsSelErrMaxOrig}{7.0}
\newcommand{\ttMsSelErrMaxOrigLang}{sq}
\newcommand{\ttMsSelErrSw}{4.8}
\newcommand{\ttMsSelErrSwOrig}{5.2}

\newcommand{\ttMsIntDelta}{0.7}

\newcommand{\ttMsIntDeltaOrig}{11.3}

\newcommand{\ttMsIntDiffDelta}{\ensuremath{-}10.6}
\newcommand{\ttMsIntDiffDeltaLo}{\ensuremath{-}11.6}
\newcommand{\ttMsIntDiffDeltaHi}{\ensuremath{-}9.6}
\newcommand{\ttPxDelta}{1.4}
\newcommand{\ttPxDeltaLo}{0.2}
\newcommand{\ttPxDeltaHi}{2.4}
\newcommand{\ttPxDeltaOrig}{8.6}
\newcommand{\ttPxDeltaOrigLo}{7.2}
\newcommand{\ttPxDeltaOrigHi}{9.9}

\newcommand{\ttPxCovEn}{78}
\newcommand{\ttPxCovSw}{76}
\newcommand{\ttPxCovJv}{76}
\newcommand{\ttPxCovSwOrig}{48}
\newcommand{\ttPxCovJvOrig}{55}
\newcommand{\ttPxCovMin}{60}
\newcommand{\ttPxCovMinLang}{km}

\newcommand{\ttPxAccEn}{88}
\newcommand{\ttPxAccMin}{73}

\newcommand{\ttPxAccMean}{85}
\newcommand{\ttPxAccMeanOrig}{82}
\newcommand{\ttPxAccMinOrig}{69}

\newcommand{\ttPxSelErrMax}{7.8}
\newcommand{\ttPxSelErrMaxLang}{km}
\newcommand{\ttPxNCovBelow}{8}

\newcommand{\ttPxNCovBelowOrig}{47}

\newcommand{\ttPxCovMinOrig}{48}
\newcommand{\ttPxCovMinOrigLang}{cy}
\newcommand{\ttPxCovGapMin}{17}
\newcommand{\ttPxCovGapMinOrig}{30}
\newcommand{\ttPxSelErrMaxOrig}{8.2}
\newcommand{\ttPxSelErrMaxOrigLang}{sq}
\newcommand{\ttPxSelErrSw}{6.1}
\newcommand{\ttPxSelErrSwOrig}{4.0}

\newcommand{\ttPxIntDelta}{0.9}

\newcommand{\ttPxIntDeltaOrig}{8.2}

\newcommand{\ttPxIntDiffDelta}{\ensuremath{-}7.3}
\newcommand{\ttPxIntDiffDeltaLo}{\ensuremath{-}8.1}
\newcommand{\ttPxIntDiffDeltaHi}{\ensuremath{-}6.5}

\newcommand{\ttRtExactShare}{18.4}
\newcommand{\ttRtExactShareMin}{8.8}
\newcommand{\ttRtExactShareMinLang}{ja}
\newcommand{\ttRtExactShareMax}{35.2}
\newcommand{\ttRtExactShareMaxLang}{af}
\newcommand{\ttRtChrfMean}{65.8}

\newcommand{\ttRtNonExactDelta}{1.1}

\newcommand{\ttRtNonExactDeltaOrig}{11.4}

\newcommand{\ttRtNonExactDiffDelta}{\ensuremath{-}10.3}
\newcommand{\ttRtNonExactDiffDeltaLo}{\ensuremath{-}11.5}
\newcommand{\ttRtNonExactDiffDeltaHi}{\ensuremath{-}9.2}

\newcommand{\ttRtLowSimDelta}{0.9}

\newcommand{\ttRtLowSimDeltaOrig}{11.5}

\newcommand{\ttRtLowSimDiffDelta}{\ensuremath{-}10.6}
\newcommand{\ttRtLowSimDiffDeltaLo}{\ensuremath{-}12.0}
\newcommand{\ttRtLowSimDiffDeltaHi}{\ensuremath{-}9.3}

\newcommand{\ttMsRtNonExactDelta}{1.1}

\newcommand{\ttMsRtNonExactDeltaOrig}{11.6}

\newcommand{\ttMsRtNonExactDiffDelta}{\ensuremath{-}10.5}
\newcommand{\ttMsRtNonExactDiffDeltaLo}{\ensuremath{-}11.5}
\newcommand{\ttMsRtNonExactDiffDeltaHi}{\ensuremath{-}9.4}

\newcommand{\ttMsRtLowSimDelta}{1.4}

\newcommand{\ttMsRtLowSimDeltaOrig}{12.6}

\newcommand{\ttMsRtLowSimDiffDelta}{\ensuremath{-}11.2}
\newcommand{\ttMsRtLowSimDiffDeltaLo}{\ensuremath{-}12.4}
\newcommand{\ttMsRtLowSimDiffDeltaHi}{\ensuremath{-}10.0}

\newcommand{\ttPxRtNonExactDelta}{1.7}

\newcommand{\ttPxRtNonExactDeltaOrig}{8.8}

\newcommand{\ttPxRtNonExactDiffDelta}{\ensuremath{-}7.1}
\newcommand{\ttPxRtNonExactDiffDeltaLo}{\ensuremath{-}7.9}
\newcommand{\ttPxRtNonExactDiffDeltaHi}{\ensuremath{-}6.3}

\newcommand{\ttPxRtLowSimDelta}{2.7}

\newcommand{\ttPxRtLowSimDeltaOrig}{10.8}

\newcommand{\ttPxRtLowSimDiffDelta}{\ensuremath{-}8.0}
\newcommand{\ttPxRtLowSimDiffDeltaLo}{\ensuremath{-}9.0}
\newcommand{\ttPxRtLowSimDiffDeltaHi}{\ensuremath{-}7.1}

\newcommand{\lorLuDeltaEn}{8.3}
\newcommand{\lorLuDeltaEnLo}{2.8}
\newcommand{\lorLuDeltaEnHi}{13.5}
\newcommand{\lorLuDeltaLoc}{18.1}
\newcommand{\lorLuDeltaLocLo}{12.0}
\newcommand{\lorLuDeltaLocHi}{23.9}

\newcommand{\lorLuDeltaDiffLo}{7.6}
\newcommand{\lorLuDeltaDiffHi}{12.0}
\newcommand{\lorMsDeltaEn}{10.7}
\newcommand{\lorMsDeltaEnLo}{7.3}
\newcommand{\lorMsDeltaEnHi}{14.0}
\newcommand{\lorMsDeltaLoc}{8.4}
\newcommand{\lorMsDeltaLocLo}{4.2}
\newcommand{\lorMsDeltaLocHi}{12.5}

\newcommand{\lorMsDeltaDiffLo}{\ensuremath{-}5.3}
\newcommand{\lorMsDeltaDiffHi}{0.6}

\newcommand{\lorMsSpearmanAuto}{0.91}

\newcommand{\lorMsSpreadEn}{41}
\newcommand{\lorMsSpreadLoc}{38}

\newcommand{\lorPxDeltaEn}{8.1}
\newcommand{\lorPxDeltaEnLo}{5.4}
\newcommand{\lorPxDeltaEnHi}{10.8}
\newcommand{\lorPxDeltaLoc}{7.4}
\newcommand{\lorPxDeltaLocLo}{4.0}
\newcommand{\lorPxDeltaLocHi}{10.7}

\newcommand{\lorPxDeltaDiffLo}{\ensuremath{-}2.2}
\newcommand{\lorPxDeltaDiffHi}{0.7}

\newcommand{\lorPxSpearmanAuto}{0.92}

\newcommand{\lorPxSpreadEn}{33}
\newcommand{\lorPxSpreadLoc}{33}

\newcommand{\budNSplits}{200}
\newcommand{\budNHeldMin}{500}

\newcommand{\budLunaEightyThreeFsCert}{0}

\newcommand{\budLunaEightyThreeFsCovEn}{0}
\newcommand{\budLunaEightyThreeFsCovMean}{0}

\newcommand{\budLunaEightyThreeOwnMiss}{61.6}

\newcommand{\budLunaOneFiftyFsCert}{2}

\newcommand{\budLunaOneFiftyFsCovEn}{11}
\newcommand{\budLunaOneFiftyFsCovMean}{0.8}

\newcommand{\budLunaOneFiftyOwnMiss}{59.1}

\newcommand{\budLunaThreeHundredFsCert}{11}

\newcommand{\budLunaThreeHundredFsCovEn}{14}
\newcommand{\budLunaThreeHundredFsCovMean}{4}

\newcommand{\budLunaThreeHundredOwnMiss}{51.4}

\newcommand{\budLunaFiveHundredFsCert}{18}

\newcommand{\budLunaFiveHundredFsCovEn}{24}
\newcommand{\budLunaFiveHundredFsCovMean}{6}

\newcommand{\budLunaFiveHundredLttCert}{0.6}

\newcommand{\budLunaFiveHundredOwnMiss}{48.9}

\newcommand{\budMsEightyThreeFsCert}{0.2}

\newcommand{\budMsEightyThreeFsCovEn}{0.6}
\newcommand{\budMsEightyThreeFsCovMean}{0.1}

\newcommand{\budMsEightyThreeOwnMiss}{57.7}

\newcommand{\budMsOneFiftyFsCert}{30}

\newcommand{\budMsOneFiftyFsCovEn}{29}
\newcommand{\budMsOneFiftyFsCovMean}{15}

\newcommand{\budMsOneFiftyOwnMiss}{56.3}

\newcommand{\budMsThreeHundredFsCert}{63}

\newcommand{\budMsThreeHundredFsCovEn}{61}
\newcommand{\budMsThreeHundredFsCovMean}{32}

\newcommand{\budMsThreeHundredOwnMiss}{52.5}

\newcommand{\budMsFiveHundredFsCert}{87}

\newcommand{\budMsFiveHundredFsCovEn}{74}
\newcommand{\budMsFiveHundredFsCovMean}{45}

\newcommand{\budMsFiveHundredLttCert}{23}

\newcommand{\budMsFiveHundredOwnMiss}{52.7}

\newcommand{\budPxEightyThreeFsCert}{4}

\newcommand{\budPxEightyThreeFsCovEn}{23}
\newcommand{\budPxEightyThreeFsCovMean}{2}

\newcommand{\budPxEightyThreeOwnMiss}{54.4}

\newcommand{\budPxOneFiftyFsCert}{27}

\newcommand{\budPxOneFiftyFsCovEn}{46}
\newcommand{\budPxOneFiftyFsCovMean}{15}

\newcommand{\budPxOneFiftyOwnMiss}{53.1}

\newcommand{\budPxThreeHundredFsCert}{55}

\newcommand{\budPxThreeHundredFsCovEn}{71}
\newcommand{\budPxThreeHundredFsCovMean}{31}

\newcommand{\budPxThreeHundredOwnMiss}{50.2}

\newcommand{\budPxFiveHundredFsCert}{78}

\newcommand{\budPxFiveHundredFsCovEn}{78}
\newcommand{\budPxFiveHundredFsCovMean}{46}

\newcommand{\budPxFiveHundredFsCertEn}{100}
\newcommand{\budPxFiveHundredFsCertLow}{12}
\newcommand{\budPxFiveHundredFsCertLowLang}{km}

\newcommand{\budPxFiveHundredLttCert}{19}

\newcommand{\budPxFiveHundredOwnMiss}{51.2}

\newcommand{\eqNHeld}{500}
\newcommand{\eqBHeld}{1{,}000}

\newcommand{\eqLunaHeldGap}{17.3}
\newcommand{\eqLunaHeldGapLo}{8.5}
\newcommand{\eqLunaHeldGapHi}{23.4}

\newcommand{\eqMsHeldGap}{18.0}
\newcommand{\eqMsHeldGapLo}{15.4}
\newcommand{\eqMsHeldGapHi}{21.9}

\newcommand{\eqPxHeldGap}{18.9}
\newcommand{\eqPxHeldGapLo}{15.4}
\newcommand{\eqPxHeldGapHi}{21.9}

\newcommand{\eqHeldGapMin}{17.3}
\newcommand{\eqHeldGapMax}{18.9}
\newcommand{\eqHeldNBelowMin}{7}
\newcommand{\eqHeldNBelowMax}{46}

\newcommand{\eqMissMin}{48.9}
\newcommand{\eqMissMax}{52.7}
\newcommand{\eqErrMin}{5.1}
\newcommand{\eqErrMax}{5.2}

\newcommand{\eqLunaFiveHundredOrigCert}{18}

\newcommand{\eqLunaFiveHundredTopCert}{10}
\newcommand{\eqLunaFiveHundredTopCovEn}{2}
\newcommand{\eqLunaFiveHundredTopCovMean}{3}

\newcommand{\eqttLunaGap}{13.7}
\newcommand{\eqttLunaGapLo}{6.7}
\newcommand{\eqttLunaGapHi}{20.3}
\newcommand{\eqttLunaDMean}{3.6}
\newcommand{\eqttLunaDMeanLo}{\ensuremath{-}3.7}
\newcommand{\eqttLunaDMeanHi}{8.7}

\newcommand{\eqttMsGap}{6.7}
\newcommand{\eqttMsGapLo}{5.0}
\newcommand{\eqttMsGapHi}{9.9}
\newcommand{\eqttMsDMean}{11.3}
\newcommand{\eqttMsDMeanLo}{9.2}
\newcommand{\eqttMsDMeanHi}{13.1}

\newcommand{\eqttPxGap}{10.8}
\newcommand{\eqttPxGapLo}{7.8}
\newcommand{\eqttPxGapHi}{12.9}
\newcommand{\eqttPxDMean}{8.1}
\newcommand{\eqttPxDMeanLo}{6.4}
\newcommand{\eqttPxDMeanHi}{9.9}

\newcommand{\eqttGapMin}{6.7}
\newcommand{\eqttGapMax}{13.7}
\newcommand{\eqttErrMin}{5.1}
\newcommand{\eqttErrMax}{5.4}

\newcommand{\injNH}{622}
\newcommand{\injNHLangs}{16}
\newcommand{\injNM}{640}
\newcommand{\injNIntents}{40}
\newcommand{\injLunaUnanswered}{15}
\newcommand{\injNReq}{7{,}644}
\newcommand{\injHGapMin}{25.0}
\newcommand{\injHGapMax}{33.4}
\newcommand{\injMGapMin}{13.3}
\newcommand{\injMGapMax}{32.1}
\newcommand{\injMLangGapMin}{5.7}
\newcommand{\injMLangGapMinLo}{2.5}
\newcommand{\injMLangGapMinHi}{8.9}
\newcommand{\injMLangGapMinLang}{sw}
\newcommand{\injMLangGapMinArm}{gpt-6-luna}

\newcommand{\xlmDelta}{1.8}
\newcommand{\xlmDeltaLo}{0.9}
\newcommand{\xlmDeltaHi}{2.7}

\newcommand{\xlmNCovBelow}{26}

\newcommand{\xlmCovEn}{91}

\newcommand{\xlmSelErrMax}{13.9}
\newcommand{\xlmSelErrMaxLang}{km}

\newcommand{\xlmEceMin}{9.4}
\newcommand{\xlmEceMax}{14.0}
\newcommand{\xlmHgcEn}{96}

\newcommand{\MsDec}{Microsoft-Decision-1}
\newcommand{\PxDec}{Perplexity Decider}

\title{One Confidence Threshold, Unequal Coverage:\\ Language Gaps in Typed Decision Interfaces\\ and How Translating Inputs Reduces Them}
\author{\name Junwei Yao \email yao.junw@northeastern.edu \\ \addr Independent Researcher
\AND
\name Yongxi Zhou \email zhou.yongx@northeastern.edu \\ \addr Independent Researcher
\AND
\name Zike Yang \email zikeyang@alumni.usc.edu \\ \addr Independent Researcher
\AND
\name Zhijun Wang \email g674901087@gmail.com \\ \addr Independent Researcher
\AND
\name Jianan Liu \email jiananliu2408@gmail.com \\ \addr Independent Researcher
\AND
\name Yanxian Chen \email chen.yanx@northeastern.edu \\ \addr Independent Researcher
\AND
\name Wenbo Ye \email ywb6113@gmail.com \\ \addr Independent Researcher}

\begin{document}
\maketitle
{\let\thefootnote\relax\footnotetext{AI assistance: The authors used Claude Code (Anthropic) during the preparation of this work.}}

\begin{abstract}
Deployed classifiers often act on a prediction only when its confidence clears a threshold and pass the rest to a
person. We ask whether one threshold serves all languages equally on typed decision interfaces, APIs that return a
probability for each option of a fixed multiple-choice question. In a confirmatory study with a written analysis plan,
three vendors' interfaces classified \cfNItems{} MASSIVE items, each in 51 languages, with a threshold of $0.9$. Even when a
request is answered correctly in both English and another language, the English version clears the threshold more
often, by \cfPxDelta--\cfMsDelta{} percentage points on average. Observed coverage, the share of
requests handled automatically, is lower than English's in every other language, while the requests the gate handles
are rarely wrong. The gap remains when every language is reweighted to the same accuracy, and a standard
calibration metric (ECE) does not detect it on one interface. In exploratory analyses, English's coverage stays higher when each
language gets its own threshold fitted to the same 5\% error target. Translating inputs into English nearly closes the
gap at $0.9$ but leaves English ahead at that error target. Requests written
natively in African languages show a gap in the same direction when English is their human translation. We recommend reporting
per-language coverage at the operating threshold.
\end{abstract}

\section{Introduction}

A common way to deploy a classifier is to let it act alone only when it is confident. The application reads the
probability of the top option, handles the request automatically when that probability clears a threshold, and
otherwise defers it to a person. This rule, which we call a confidence gate, is selective prediction with a reject
option \citep{chow1970reject,geifman2017selective}. Typed decision interfaces, a kind of model API first offered with TypeSafe AI's Jev \citep{typesafe2026jev}
and since by OpenAI, Microsoft and Perplexity \citep{openai2026decisions,microsoft2026decision1,perplexity2026decider}, make the rule easy to apply. The developer writes a question with a
fixed set of options, and the model returns one probability per option for an input such as a user's message.
``Typed'' refers to the question's declared answer type, such as Boolean or Choice. A developer whose users write in
many languages can apply one threshold to all of them. We ask how evenly that threshold serves the languages, whether
the difference survives changes of accuracy, schema and threshold, and what reduces it.

\begin{figure}[t]
\centering\small
\fbox{\begin{minipage}{0.97\linewidth}
\textbf{Question} (Choice, written in English): ``What does the user want the assistant to do?''\\
\textbf{Options}: the 59 MASSIVE intents, such as \texttt{alarm\_set}, \texttt{datetime\_query} and
\texttt{play\_music}.\\[4pt]
\centering
\begin{tabular}{@{}lllcl@{}}
\toprule
Language & Input (MASSIVE's version) & Top option & Probability & Gate at $0.9$\\
\midrule
English & \emph{\exUttEn} & \exGold{} (correct) & \exPEn & handled automatically\\
Swahili & \emph{\exUttSw} & \exGold{} (correct) & \exPSw & deferred to a person\\
Javanese & \emph{\exUttJv} & \exGold{} (correct) & \exPJv & deferred to a person\\
\bottomrule
\end{tabular}
\end{minipage}}
\caption{One MASSIVE item in three of its \MlNLangs{} language versions, sent to gpt-6-luna. The model picks the
correct intent in all three, but only the English probability clears the threshold, so the gate defers the correct
Swahili and Javanese answers to a person.}
\label{fig:example}
\end{figure}

Figure~\ref{fig:example} shows the effect on one item. We measure it with two standard quantities of selective
prediction: coverage, the share of requests the gate handles automatically, and selective error, the error rate among
those it handles. Earlier multilingual work asks whether the probabilities are trustworthy, which is calibration
\citep{arxiv2210_12265,arxiv2311_08669_ax,krause2023confidently,huang2026multilingualcalib}, or how accurate a
model's kept answers are when it may abstain \citep{feng2024abstain_tc}, or how well confidence ranks errors in each
language \citep{arxiv2610_14112_ax}. A model can be well calibrated in two languages and still be less confident in one
of them. That lower confidence can be what calibration requires in a language the model answers less accurately, so we
also fit each language its own threshold for the same 5\% selective error and ask whether English still covers more.

Every language receives the same items, in the datasets' translations. The main evidence is a confirmatory study on
\cfNItems{} MASSIVE intent items \citep{arxiv2204_08582_ax} in each of \MlNLangs{} languages, sent to three typed
decision interfaces from different vendors. Exploratory analyses on these items and on a separate set of \MlNPerLang{}
MASSIVE and \XnNPerLang{} XNLI items \citep{conneau2018xnli_tc} per language compare the gap with other classifiers,
schemas and formats and test what reduces it. Both datasets translate English originals, so a further exploratory
check uses requests written natively in \injNHLangs{} African languages \citep{yu2025injongo}, with human or machine
English translations.

\paragraph{Findings.} The first finding is the confirmatory result; the others are exploratory.
\begin{itemize}\setlength{\itemsep}{1pt}
\item \emph{The gap.} In all three interfaces, among items answered correctly in both English and another language,
  the gate handles English more often, by \cfPxDelta--\cfMsDelta{} points on average. Observed coverage is lower than
  English's in every other language, detectably in \cfMsNCovBelow--\cfPxNCovBelow, while the requests the gate handles are
  rarely wrong (Section~\ref{sec:res}).
\item \emph{A common error target.} With each language's threshold fitted to 5\% selective error on tuning items,
  English's held-out coverage still exceeds the mean other language's by \eqHeldGapMin--\eqHeldGapMax{} points
  (Section~\ref{sec:res}).
\item \emph{Comparisons.} The gap remains after standardizing every language to a 75\% share of correct answers and
  with a schema written in each input's language. A classifier fine-tuned on every MASSIVE language has a much smaller
  gap, at a higher error (Section~\ref{sec:rob}).
\item \emph{Natively written inputs.} When the non-English request is the original and English its human
  translation, the English version is still handled more often, by \injHGapMin--\injHGapMax{} points
  (Section~\ref{sec:rob}).
\item \emph{What reduces it.} Translating each input into English before the gate reduces the gap at $0.9$ to
  \ttDelta--\ttPxDelta{} points; at a common 5\% error target it narrows English's lead on two interfaces
  (Section~\ref{sec:thr}).
\item \emph{What does not.} Thresholds fitted on each language's own labels leave English's coverage higher, and
  per-language conformal sets widen English's lead in single-label sets on gpt-6-luna. A fixed-sequence test certifies per-language
  thresholds in more than half of language--split pairs from 300 labels per language upward, and only on two interfaces
  (Section~\ref{sec:thr}).
\end{itemize}

\section{Setup}\label{sec:setup}

\paragraph{Interface and gate.} A request carries an input and a Choice question about it, and the response gives one
probability per option. For MASSIVE the input is the user's utterance and the question asks for the user's intent,
with the 59 intents of the test split as options and each intent's name (``alarm set'') as its description. For XNLI
the input is a premise--hypothesis pair and the question asks how the hypothesis relates to the premise, with three
options and one-sentence descriptions. We call the question text and the option descriptions the schema; option values
such as \texttt{alarm\_set} are not part of it. The schema stays in English while the input changes language, as
when a developer writes the schema once for users of all languages. At threshold
$t$ the gate handles a request automatically when the top option's probability is at least $t$ and defers it
otherwise. The returned option is always the top option, and confidence means its probability, not the separate
\texttt{confidence} field that the interfaces also return. The main threshold is $0.9$.

\paragraph{Models.} gpt-6-luna is served by OpenAI's Decisions API \citep{openai2026decisions}, a public beta whose model
name carries no snapshot version; it returned probabilities rounded to two decimals. \MsDec{} \citep{microsoft2026decision1} and \PxDec{} v1.1 27B \citep{perplexity2026decider} are
served by OpenRouter's decision endpoint with provider fallback disabled, and every response named the same snapshot
(\texttt{microsoft-decision-1-20261009} and \texttt{pplx-decider-v1.1-27b-20261006}). They were chosen as the
OpenRouter decision models from vendors other than OpenAI that have a versioned name and return full-precision
probabilities over all 59 options (Appendix~\ref{app:confirm}).

\paragraph{Data.} An item is one source utterance or sentence pair with all its language versions; a request is one
language version of one item. The confirmatory study draws \cfNItems{} MASSIVE test items at random from those present
in all \MlNLangs{} languages, excluding every item of the exploratory study and an earlier pilot, so each model receives \cfNReq{} requests. The exploratory study
uses \MlNPerLang{} other random MASSIVE items and \XnNPerLang{} XNLI test pairs in all \XnNLangs{} XNLI languages. Tables name languages by their ISO 639-1 codes (en
English, sw Swahili, jv Javanese, ur Urdu; zh-CN and zh-TW are simplified and traditional Chinese in MASSIVE, and zh is Chinese in XNLI).

\paragraph{Measures.} Per language we report coverage, selective error, accuracy of the top option, and top-label
expected calibration error (ECE) over 10 equal-width bins. Our main measure compares like with like. Take the set
$B_L$ of items whose top option is correct in both English and language $L$. On these items, $d_L$ is how much more
often the gate handles the English version, and the mean paired handling gap $\Delta$ averages $d_L$ over the 50
other languages:
\[
d_L = \frac{1}{|B_L|} \sum_{u \in B_L} \bigl(H_{\mathrm{en}}(u) - H_L(u)\bigr), \qquad
\Delta = \frac{1}{50} \sum_{L \ne \mathrm{en}} d_L,
\]
where $H(u)$ is 1 if the gate handles that version of item $u$ and 0 otherwise.

\paragraph{Refusals.} A refused or unanswered request counts as deferred and wrong, and is left out of ECE.
gpt-6-luna refused \MlNRefused{} of \MlNRequests{} exploratory MASSIVE requests and \XnNRefused{} of \XnNRequests{}
XNLI requests and answered \cfLunaAnswered\% of confirmatory requests and \ttAnswered\% of translated ones; the other
two interfaces answered every request.

\paragraph{Confirmatory plan.} The plan fixed the item sample, the three models, the threshold, the scoring rules
above and $\Delta$ as the primary endpoint, one test per model. Inference resamples the \cfNItems{} items
\cfB{} times, each resample shared by all languages, and gives a percentile 95\% interval and a two-sided bootstrap
$p$-value; Holm's procedure adjusts the three $p$-values. The confirmatory success criterion for a model is an
adjusted $p$ below $0.05$ with $\Delta > 0$. Seven secondary measures, reported in Sections~\ref{sec:res}
and~\ref{sec:rob} and Appendix~\ref{app:confirm}, carry no decision rule. The plan was written before the first confirmatory request, and every planned request was sent
(Appendix~\ref{app:confirm}).

\paragraph{Comparing languages with English.} Testing 50 languages against English one at a time would flag some
differences by chance. We therefore use a simultaneous 95\% band: a bootstrap interval for each language's difference
from English, widened so that all 50 intervals hold together with 95\% probability (Appendix~\ref{app:repro}). A
difference is detected when its band excludes zero.

\paragraph{Exploratory analyses.} Every analysis of the \MlNPerLang-item MASSIVE and the XNLI data, and every
analysis of the confirmatory items outside the plan, is exploratory. Corrections within an analysis are stated with
it; we do not correct across exploratory analyses.

\section{Does one threshold serve the languages equally?}\label{sec:res}

\begin{table}[t]
\centering\small
\caption{In all three interfaces, the gate handles English more often, including on items answered correctly in both
languages. Confirmatory study, gate at $0.9$. $\Delta$: mean paired handling gap with 95\% bootstrap interval.
``simult.'': languages whose difference from English the simultaneous band detects. Correct, deferred: share of all
requests deferred although the top label was correct. Standardized gap: English minus mean language in coverage at a
$0.75$ share of correct answers, with 95\% interval. A language named at a minimum or maximum is chosen before
rounding. In \% or points.}\label{tab:confirm}
\setlength{\tabcolsep}{4pt}
\begin{tabular}{@{}lccc@{}}
\toprule
 & gpt-6-luna & \MsDec & \PxDec \\
\midrule
$\Delta$, primary endpoint & \cfLunaDelta{} [\cfLunaDeltaLo, \cfLunaDeltaHi] & \cfMsDelta{} [\cfMsDeltaLo, \cfMsDeltaHi] & \cfPxDelta{} [\cfPxDeltaLo, \cfPxDeltaHi] \\
Holm-adjusted $p$ & $\cfLunaPHolm$ & $\cfMsPHolm$ & $\cfPxPHolm$ \\
Languages with $d_L > 0$ (of 50, descriptive) & \cfLunaNPdPos & \cfMsNPdPos & \cfPxNPdPos \\
Smallest / largest $d_L$ & \cfLunaPdMin{} (\cfLunaPdMinLang) / \cfLunaPdMax{} (\cfLunaPdMaxLang) & \cfMsPdMin{} (\cfMsPdMinLang) / \cfMsPdMax{} (\cfMsPdMaxLang) & \cfPxPdMin{} (\cfPxPdMinLang) / \cfPxPdMax{} (\cfPxPdMaxLang) \\
\midrule
Coverage, en / sw / jv & \cfLunaCovEn{} / \cfLunaCovSw{} / \cfLunaCovJv & \cfMsCovEn{} / \cfMsCovSw{} / \cfMsCovJv & \cfPxCovEn{} / \cfPxCovSw{} / \cfPxCovJv \\
Lowest coverage & \cfLunaCovMin{} (\cfLunaCovMinLang) & \cfMsCovMin{} (\cfMsCovMinLang) & \cfPxCovMin{} (\cfPxCovMinLang) \\
Coverage below / above en (simult.) & \cfLunaNCovBelow{} / \cfLunaNCovAbove & \cfMsNCovBelow{} / \cfMsNCovAbove & \cfPxNCovBelow{} / \cfPxNCovAbove \\
Accuracy, en / lowest & \cfLunaAccEn{} / \cfLunaAccMin & \cfMsAccEn{} / \cfMsAccMin & \cfPxAccEn{} / \cfPxAccMin \\
Selective error, en & \cfLunaSelErrEn & \cfMsSelErrEn & \cfPxSelErrEn \\
Highest selective error & \cfLunaSelErrMax{} (\cfLunaSelErrMaxLang) & \cfMsSelErrMax{} (\cfMsSelErrMaxLang) & \cfPxSelErrMax{} (\cfPxSelErrMaxLang) \\
\midrule
ECE, en / range & \cfLunaEceEn{} / \cfLunaEceMin--\cfLunaEceMax & \cfMsEceEn{} / \cfMsEceMin--\cfMsEceMax & \cfPxEceEn{} / \cfPxEceMin--\cfPxEceMax \\
ECE above / below en (simult.) & \cfLunaNEceWorse{} / \cfLunaNEceBetter & \cfMsNEceWorse{} / \cfMsNEceBetter & \cfPxNEceWorse{} / \cfPxNEceBetter \\
\midrule
Handled among correct, en / sw & \cfLunaHgcEn{} / \cfLunaHgcSw & \cfMsHgcEn{} / \cfMsHgcSw & \cfPxHgcEn{} / \cfPxHgcSw \\
Handled among correct below en (simult.) & \cfLunaNHgcBelow & \cfMsNHgcBelow & \cfPxNHgcBelow \\
Correct, deferred, en / sw & \cfLunaCdEn{} / \cfLunaCdSw & \cfMsCdEn{} / \cfMsCdSw & \cfPxCdEn{} / \cfPxCdSw \\
Standardized gap & \cfLunaStd{} [\cfLunaStdLo, \cfLunaStdHi] & \cfMsStd{} [\cfMsStdLo, \cfMsStdHi] & \cfPxStd{} [\cfPxStdLo, \cfPxStdHi] \\
\bottomrule
\end{tabular}
\end{table}

\paragraph{The mean paired handling gap favors English in all three interfaces.} The primary endpoint meets the
confirmatory success criterion in all three models (Table~\ref{tab:confirm}): on items answered correctly in both languages, the
gate handles the English version more often by \cfPxDelta--\cfMsDelta{} points on average over the 50 other
languages (Holm-adjusted $p<0.001$ for each). All 150 per-language estimates $d_L$ are positive
(Table~\ref{tab:confirmlang}, Appendix~\ref{app:confirm}).

\paragraph{Observed coverage is lower than English's in all 50 languages.} With one threshold, gpt-6-luna
handles \cfLunaCovEn\% of English requests automatically, \cfLunaCovSw\% of Swahili and \cfLunaCovJv\% of Javanese,
the lowest. The other two interfaces are more confident overall but show the same pattern. Observed coverage is below
English's in all 50 languages for each model, detectably so in \cfMsNCovBelow--\cfPxNCovBelow{} of them. The handled
requests are almost always correct: the largest observed selective error is \cfPxSelErrMax\% (Table~\ref{tab:confirm}).

\paragraph{For gpt-6-luna, ECE does not show the gap.} Each language's observed ECE differs from English's by at
most \cfLunaEceGapMaxAbs{} points, and the simultaneous band (median half-width \cfLunaEceHalfMed{} points) detects no
difference, while coverage trails English by up to \cfLunaCovGapMax{} points. For \MsDec{} and \PxDec{} the band
detects higher ECE than English's in \cfMsNEceWorse{} and \cfPxNEceWorse{} languages and lower in none, but coverage
differences in many more.

\paragraph{Much of the gap is correct answers sent to a person.} Of the correctly answered requests, gpt-6-luna's gate
handles \cfLunaHgcEn\% in English and \cfLunaHgcSw\% in Swahili; this share is below English's in all 50 languages
for each model, detectably so in \cfLunaNHgcBelow{} for gpt-6-luna. The gate defers \cfLunaCdSw\% of all Swahili
requests although the top label was correct, against \cfLunaCdEn\% of English requests. This share follows from
three standard quantities:
\[
P(\text{correct and deferred}) = \text{accuracy} - \text{coverage} \times (1 - \text{selective error}).
\]

\begin{table}[t]
\centering\small
\caption{The exploratory study on MASSIVE: coverage varies across languages while observed selective error stays
between \MlAutoErrLo\% and \MlAutoErrHi\%. gpt-6-luna, selected languages (all in Appendix~\ref{app:full}), in \%.
acc: accuracy. auto, err: coverage and selective error at $0.9$. hand/both: among items correct in both English and
the language, the share handled in English but deferred in the language, and the share deferred in both. EN thr, own
thr (Section~\ref{sec:thr}): held-out coverage/selective error under a threshold fitted to a 5\% error budget on
English's or the language's own tuning items. ``--'': no threshold meets the budget. $\dagger$: error above 5\%,
also when it rounds to 5.}\label{tab:main}
\begin{tabular}{lrrrrrcc}
\toprule
 & & & & & paired & EN thr & own thr \\
lang & acc & ECE & auto & err & hand/both & cov/err & cov/err \\
\midrule
en & 83 & 5 & 60 & 1 & 0 / 28 & 72 / 0 & 72 / 0 \\
nl & 83 & 8 & 53 & 0 & 14 / 19 & 73 / 2 & 87 / 5$\dagger$ \\
de & 78 & 4 & 58 & 3 & 10 / 18 & 76 / 2 & 42 / 0 \\
zh-CN & 79 & 5 & 55 & 1 & 14 / 16 & 67 / 2 & 72 / 4 \\
ja & 83 & 6 & 54 & 1 & 16 / 15 & 72 / 0 & 73 / 0 \\
hi & 78 & 5 & 54 & 1 & 11 / 18 & 75 / 6$\dagger$ & 78 / 6$\dagger$ \\
ar & 74 & 6 & 44 & 2 & 20 / 20 & 60 / 2 & 69 / 4 \\
ur & 76 & 9 & 46 & 1 & 17 / 21 & 64 / 2 & 64 / 2 \\
am & 69 & 7 & 40 & 0 & 25 / 17 & 58 / 5$\dagger$ & 63 / 5 \\
jv & 64 & 11 & 24 & 3 & 49 / 13 & 42 / 7$\dagger$ & 45 / 7$\dagger$ \\
sw & 68 & 7 & 29 & 0 & 39 / 17 & 46 / 3 & 37 / 0 \\
cy & 70 & 9 & 37 & 0 & 30 / 17 & 57 / 8$\dagger$ & 51 / 3 \\
\bottomrule
\end{tabular}

\end{table}

\paragraph{Lower coverage outside English appears at every tested threshold.} In the exploratory study the
simultaneous band puts \phSeventyNBelowSim, \phEightyNBelowSim, \phNinetyNBelowSim{} and \phNinetyFiveNBelowSim{}
MASSIVE languages below English at thresholds $0.7$, $0.8$, $0.9$ and $0.95$, and none above
(Appendix~\ref{app:posthoc}, Figure~\ref{fig:sweep}).

\paragraph{On XNLI, calibration also differs by language.} Selective error rises to \XnAutoErrHi\%
(\XnAutoErrHiLang), and the band detects an ECE difference against English for \XnNEceDiffResolvedSim{} of \XnNEceDiffLangs{} languages, in line with
the miscalibration outside English that \citet{arxiv2311_08669_ax} report for multilingual question answering.

\begin{figure}[t]
\centering
\includegraphics[width=0.9\linewidth]{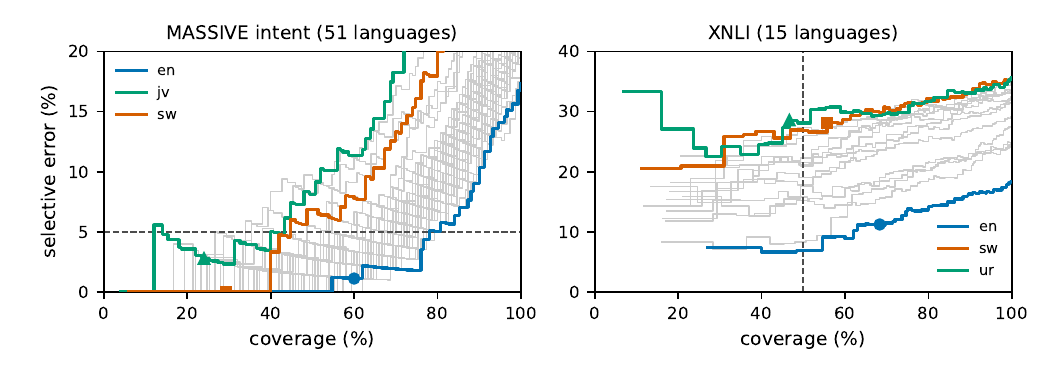}
\caption{On MASSIVE, English's empirical curve reaches the highest coverage at 5\% selective error, tied with Turkish.
gpt-6-luna, exploratory study: observed selective error against coverage as the threshold falls, one gray line per
language; markers show the gate at $0.9$. Dashed: 5\% selective error (MASSIVE), 50\% coverage (XNLI).}
\label{fig:riskcov}
\end{figure}

\paragraph{At a common error target, English still covers more.} Part of the gap arises because $0.9$ sits at a
different point on each language's risk--coverage curve: on the exploratory items, the threshold that keeps observed
selective error within 5\% ranges from \rcvMlFiveThrLo{} to \rcvMlFiveThrHi{} across MASSIVE languages (English
\rcvMlFiveThrEn). But the curves also differ (Figure~\ref{fig:riskcov}). On the confirmatory items we fit each
language's threshold to 5\% selective error on a random half and apply it to the other half, over 200 random splits (Table~\ref{tab:eqrisk},
Appendix~\ref{app:eqrisk}). English's held-out coverage then exceeds the mean other language's by
\eqHeldGapMin--\eqHeldGapMax{} points in the three interfaces, each 95\% interval above zero, at held-out selective
error of \eqErrMin--\eqErrMax\%.

\begin{table}[t]
\centering\small
\caption{With thresholds fitted to 5\% selective error, English's coverage exceeds the mean other language's, in-sample and held-out.
Confirmatory items. en, mean: coverage for English and the mean of the other 50 languages; en $-$ mean with 95\%
bootstrap interval; below en: languages whose coverage is below English's, as point estimates / detected by the
simultaneous band; lowest: lowest coverage, with ties listed (0: no threshold meets 5\%); err: held-out selective error, averaged over
splits and then languages; miss: share of language--split pairs whose held-out
selective error exceeds 5\%. In \% or points.}\label{tab:eqrisk}
\setlength{\tabcolsep}{4pt}\begin{tabular}{@{}llrrlrlrr@{}}
\toprule
interface & threshold & en & mean & en $-$ mean [95\%] & below en & lowest (lang) & err & miss \\
\midrule
gpt-6-luna & in-sample & 56.9 & 40.9 & 16.0 [4.9, 27.7] & 49 / 1 & 0.0 (ar) & -- & -- \\
 & held-out & 57.8 & 40.5 & 17.3 [8.5, 23.4] & 50 / 7 & 9.8 (jv) & 5.2 & 48.9 \\
\MsDec & in-sample & 80.4 & 63.0 & 17.4 [14.3, 22.8] & 50 / 21 & 28.5 (my) & -- & -- \\
 & held-out & 81.1 & 63.1 & 18.0 [15.4, 21.9] & 50 / 30 & 29.4 (my) & 5.2 & 52.7 \\
\PxDec & in-sample & 86.9 & 67.9 & 19.0 [14.6, 22.5] & 50 / 31 & 40.5 (km) & -- & -- \\
 & held-out & 86.7 & 67.8 & 18.9 [15.4, 21.9] & 50 / 46 & 39.6 (km) & 5.1 & 51.2 \\
\bottomrule
\end{tabular}

\end{table}

\section{Diagnostic comparisons}\label{sec:rob}

The standardization and the classifier use the confirmatory items, the native-input check uses InjongoIntent, and the
schema, format and open-weight checks use the exploratory items, with gpt-6-luna unless noted.

\paragraph{The gap remains after standardizing every language to a 75\% share of correct answers.} Most languages
are less accurate than English, so some coverage gap is expected. This planned secondary measure recomputes each
language's coverage as if $0.75$ of its answers were correct, keeping its own handling rates among correct and among
wrong answers. English still leads the mean language by \cfPxStd--\cfMsStd{} points, with every 95\% interval above
zero (Table~\ref{tab:confirm}).

\begin{table}[t]
\centering\small
\caption{The English version is handled more often also when it is the translation. InjongoIntent requests written
natively in African languages, gate at $0.9$, exploratory. Human pairs: \injNH{} native utterances in \injNHLangs{}
languages and their human English translations. Machine pairs: \injNM{} native utterances in each of sw, am, ha, yo
and zu and their \ttMTModel{} English translations. cov, acc: native / English. $\Delta$: mean paired handling gap,
English minus native, on items correct in both versions, with 95\% interval; for human pairs, pooled over the
\injNH{} items; for machine pairs, the mean of the five languages' gaps. In \% or points.}\label{tab:injongo}
\setlength{\tabcolsep}{4pt}\begin{tabular}{@{}lccc c ccc@{}}
\toprule
 & \multicolumn{3}{c}{human pairs} && \multicolumn{3}{c}{machine pairs} \\
\cmidrule{2-4}\cmidrule{6-8}
 & cov & acc & $\Delta$ && cov & acc & $\Delta$ \\
\midrule
gpt-6-luna & 41 / 75 & 67 / 86 & 28.5 [23.5, 33.5] && 60 / 78 & 84 / 89 & 17.6 [15.8, 19.3] \\
\MsDec & 33 / 80 & 53 / 88 & 33.4 [27.8, 38.9] && 43 / 81 & 70 / 92 & 32.1 [29.9, 34.3] \\
\PxDec & 48 / 84 & 69 / 89 & 25.0 [20.4, 29.6] && 66 / 85 & 81 / 91 & 13.3 [11.6, 14.9] \\
\bottomrule
\end{tabular}

\end{table}

\paragraph{With natively written requests, English is handled more often even as the translation.}\label{sec:injongo}
In MASSIVE and XNLI, English is the original and every other version a translation. InjongoIntent
\citep{yu2025injongo} reverses this: native speakers of \injNHLangs{} African languages wrote its requests, covering
\injNIntents{} intents, and \injNH{} of them come with a human English translation. We sent both versions to the three
interfaces, with the MASSIVE question and these intents as options (Table~\ref{tab:injongo};
Appendix~\ref{app:newmethods}). On items correct in both versions, the gate handles the English translation more often
than the native original, by \injHGapMin--\injHGapMax{} points. Machine translation into English points the same way:
over \injNM{} native requests in each of five languages, the mean gap is \injMGapMin--\injMGapMax{} points, and every
per-language interval lies above zero (smallest \injMLangGapMin{} [\injMLangGapMinLo, \injMLangGapMinHi],
\injMLangGapMinLang{} on \injMLangGapMinArm).

\paragraph{With a machine-translated schema in the input's language, the gate still handles English more often.}\label{sec:loc}
So far every non-English input came with an English schema. We translated the question and the 59 intent descriptions
into each language with LLMs, mostly Qwen models \citep{alibaba2026qwen37flash,qwen2026qwen36_27b,qwen2026qwen38}, and
sent the same \MlNPerLang{} items per language with the schema in the input's language; option values stay English
identifiers (Appendix~\ref{app:schema}). The mean paired handling gap stays above zero in all three interfaces. On
gpt-6-luna it grows from \lorLuDeltaEn{} to \lorLuDeltaLoc{} points, a detected increase, and the localized schema
reorders the languages; on \MsDec{} and \PxDec{} it changes by less than three points, a change that is not detected
(Appendix~\ref{app:schema}).

\paragraph{A classifier fine-tuned on every language has a smaller gap and a higher error.} We gated a public
XLM-RoBERTa base model \citep{conneau2020xlmr}, fine-tuned on MASSIVE's training data in all its languages
\citep{haryoaw2024massivexlmr}, on the same confirmatory inputs (Appendix~\ref{app:newmethods}). Its mean paired
handling gap is \xlmDelta{} points [\xlmDeltaLo, \xlmDeltaHi], against \cfPxDelta--\cfMsDelta{} for the interfaces.
It handles \xlmCovEn\% of English requests, against \cfLunaCovEn--\cfPxCovEn\% for the interfaces, and its handled
requests are wrong up to \xlmSelErrMax\% of the time (\xlmSelErrMaxLang), against at most \cfPxSelErrMax\%. The
classifier also differs in training and architecture, so the comparison describes two kinds of system.

\paragraph{Wording, format and model checks.} An English rewording of the schema with the same meanings keeps
gpt-6-luna's coverage ranking (Spearman \pphRhoPara), and a corrected Swahili schema changes Swahili coverage by
\swfixDAutoOrig{} points [\swfixDAutoOrigLo, \swfixDAutoOrigHi] (Appendix~\ref{app:schema}). Asking the gate's question
together with two other questions about the same input leaves per-language coverage nearly unchanged (Spearman
\pcoMlCmpRhoCov; Appendix~\ref{app:choiceonly}). In three open-weight models of the class, English ranks first or
second in coverage at $0.9$ (Appendix~\ref{app:open}). A gate can also fail with high coverage everywhere: a small open
model, laya-multilingual \citep{convai2026layaml}, handles \LyXnAutoLo--\LyXnAutoHi\% of XNLI requests in every
language at selective error up to \LyXnAutoErrHi\%, a failure a coverage check alone would miss.

\section{What reduces the gap, and what does not}
\label{sec:thr}

\begin{table}[t]
\centering\small
\caption{Translating each input into English before the gate reduces the mean paired handling gap in all three
interfaces. Exploratory analysis of the confirmatory items, gate at $0.9$. orig: the MASSIVE input; TT: its English translation. $\Delta$: mean
paired handling gap with 95\% interval; on the fixed set: on items correct in both languages under both conditions.
Accuracy: English / mean of the other 50 languages / lowest language. Below en: languages the simultaneous band puts
below English. en $-$ lowest is computed before rounding, so it can
differ by one from the difference of the rounded values. en $-$ mean at 5\% target: English's held-out coverage
minus the mean other language's with each language's threshold fitted to 5\% selective error (as in
Table~\ref{tab:eqrisk}), with 95\% interval. In \% or points.}\label{tab:tt}
\setlength{\tabcolsep}{3pt}\footnotesize
\begin{tabular}{@{}lcccccc@{}}
\toprule
 & \multicolumn{2}{c}{gpt-6-luna} & \multicolumn{2}{c}{\MsDec} & \multicolumn{2}{c}{\PxDec} \\
\cmidrule(lr){2-3}\cmidrule(lr){4-5}\cmidrule(lr){6-7}
 & orig & TT & orig & TT & orig & TT \\
\midrule
$\Delta$ & \ttDeltaOrig{} [\ttDeltaOrigLo, \ttDeltaOrigHi] & \ttDelta{} [\ttDeltaLo, \ttDeltaHi] & \ttMsDeltaOrig{} [\ttMsDeltaOrigLo, \ttMsDeltaOrigHi] & \ttMsDelta{} [\ttMsDeltaLo, \ttMsDeltaHi] & \ttPxDeltaOrig{} [\ttPxDeltaOrigLo, \ttPxDeltaOrigHi] & \ttPxDelta{} [\ttPxDeltaLo, \ttPxDeltaHi] \\
$\Delta$ on the fixed set & \ttIntDeltaOrig & \ttIntDelta & \ttMsIntDeltaOrig & \ttMsIntDelta & \ttPxIntDeltaOrig & \ttPxIntDelta \\
Accuracy & \ttAccEn{} / \ttAccMeanOrig{} / \ttAccMinOrig & \ttAccEn{} / \ttAccMean{} / \ttAccMin & \ttMsAccEn{} / \ttMsAccMeanOrig{} / \ttMsAccMinOrig & \ttMsAccEn{} / \ttMsAccMean{} / \ttMsAccMin & \ttPxAccEn{} / \ttPxAccMeanOrig{} / \ttPxAccMinOrig & \ttPxAccEn{} / \ttPxAccMean{} / \ttPxAccMin \\
Below en (of 50) & \ttNCovBelowOrig & \ttNCovBelow & \ttMsNCovBelowOrig & \ttMsNCovBelow & \ttPxNCovBelowOrig & \ttPxNCovBelow \\
Coverage, en / sw / jv & \cfLunaCovEn{} / \ttCovSwOrig{} / \ttCovJvOrig & \ttCovEn{} / \ttCovSw{} / \ttCovJv & \ttMsCovEn{} / \ttMsCovSwOrig{} / \ttMsCovJvOrig & \ttMsCovEn{} / \ttMsCovSw{} / \ttMsCovJv & \ttPxCovEn{} / \ttPxCovSwOrig{} / \ttPxCovJvOrig & \ttPxCovEn{} / \ttPxCovSw{} / \ttPxCovJv \\
Lowest coverage & \ttCovMinOrig{} (\ttCovMinOrigLang) & \ttCovMin{} (\ttCovMinLang) & \ttMsCovMinOrig{} (\ttMsCovMinOrigLang) & \ttMsCovMin{} (\ttMsCovMinLang) & \ttPxCovMinOrig{} (\ttPxCovMinOrigLang) & \ttPxCovMin{} (\ttPxCovMinLang) \\
en $-$ lowest & \ttCovGapMinOrig & \ttCovGapMin & \ttMsCovGapMinOrig & \ttMsCovGapMin & \ttPxCovGapMinOrig & \ttPxCovGapMin \\
Selective error, sw & \ttSelErrSwOrig & \ttSelErrSw & \ttMsSelErrSwOrig & \ttMsSelErrSw & \ttPxSelErrSwOrig & \ttPxSelErrSw \\
Highest selective error & \ttSelErrMaxOrig{} (\ttSelErrMaxOrigLang) & \ttSelErrMax{} (\ttSelErrMaxLang) & \ttMsSelErrMaxOrig{} (\ttMsSelErrMaxOrigLang) & \ttMsSelErrMax{} (\ttMsSelErrMaxLang) & \ttPxSelErrMaxOrig{} (\ttPxSelErrMaxOrigLang) & \ttPxSelErrMax{} (\ttPxSelErrMaxLang) \\
\midrule
en $-$ mean at 5\% target & \eqLunaHeldGap{} [\eqLunaHeldGapLo, \eqLunaHeldGapHi] & \eqttLunaGap{} [\eqttLunaGapLo, \eqttLunaGapHi] & \eqMsHeldGap{} [\eqMsHeldGapLo, \eqMsHeldGapHi] & \eqttMsGap{} [\eqttMsGapLo, \eqttMsGapHi] & \eqPxHeldGap{} [\eqPxHeldGapLo, \eqPxHeldGapHi] & \eqttPxGap{} [\eqttPxGapLo, \eqttPxGapHi] \\
\bottomrule
\end{tabular}
\end{table}

\subsection{Translating inputs into English}
\paragraph{Translating each input into English reduces the gap in all three interfaces.} A deployer can translate each
request into English before the gate (translate-test). We translated the \ttNUtt{} non-English confirmatory inputs with \ttMTModel{}
\citep{openai2026gpt54mini} and sent them with the unchanged English schema (Appendix~\ref{app:newmethods}). The mean
paired handling gap falls from \cfPxDelta--\cfMsDelta{} to \ttDelta--\ttPxDelta{} points (Table~\ref{tab:tt}), fewer
languages fall detectably below English, and English's lead over the lowest language shrinks. A small gap remains
detectable only for \PxDec. Accuracy rises too, and so does selective error in some languages whose coverage rises,
such as Swahili on gpt-6-luna and \PxDec. Khmer has the lowest coverage after translation in all three interfaces.

Translation changes which answers are correct, so we also compare the gap on items
correct in both languages under both conditions. The MASSIVE versions were translated from English, so translating
back might simply recover the English utterance; \ttRtExactShare\% of translations match it after normalizing case,
punctuation and whitespace. The reduction stays detectable on the fixed items, without the exact matches, and on the
half of pairs least similar to the English utterance, in all three interfaces (Table~\ref{tab:ttcheck},
Appendix~\ref{app:newmethods}).

\paragraph{At a common error target, translation narrows English's lead on two interfaces.} We repeat the held-out analysis of
Table~\ref{tab:eqrisk} on the translated inputs, fitting each language's threshold to 5\% selective error on the same
tuning halves and applying it to the same held-out halves; held-out selective error is \eqttErrMin--\eqttErrMax\%.
English's lead is \eqttGapMin--\eqttGapMax{} points after translation, against \eqHeldGapMin--\eqHeldGapMax{} before
(Table~\ref{tab:tt}). The mean other language's held-out coverage rises by \eqttMsDMean{} [\eqttMsDMeanLo,
\eqttMsDMeanHi] points on \MsDec{} and \eqttPxDMean{} [\eqttPxDMeanLo, \eqttPxDMeanHi] on \PxDec; on gpt-6-luna the
change, \eqttLunaDMean{} [\eqttLunaDMeanLo, \eqttLunaDMeanHi], is not detected.

\subsection{Thresholds set per language from about a hundred labels}
A deployer can instead set thresholds from a small labeled sample, aiming at selective error of at most 5\% (the
error budget). We simulate this on the exploratory items, split by even or odd id into \MlNTune{} tuning and
\MlNTest{} held-out items per language (\XnNTune/\XnNTest{} on XNLI), and compare three thresholds:
\begin{itemize}\setlength{\itemsep}{1pt}
\item \emph{English threshold}: the lowest threshold whose selective error on English's tuning items is within budget,
  applied to every language (\phMlEnFitThr{} on MASSIVE, \phXnEnFitThr{} on XNLI).
\item \emph{Own threshold}: the same rule, fitted on each language's own tuning items.
\item \emph{Certified threshold}: Learn-then-Test \citep{angelopoulos2025ltt}, in two variants. Each of
  \MlLttGridN{} thresholds from 0.50 to 0.99 gets an exact one-sided binomial test of its tuning error. The
  \emph{Bonferroni} variant corrects over all of them; the \emph{fixed-sequence} variant tests from 0.99 downward and
  stops at the first failure. If tuning and future items are i.i.d., either gives a threshold whose population selective
  error is at most 5\% with probability at least $0.9$. If none passes, the gate defers everything.
\end{itemize}

\paragraph{The English threshold misses the budget in many languages.} On the held-out items it exceeds the 5\%
budget in \MlNEnThrOverTarget{} of \MlNLangs{} MASSIVE languages and twice the budget in \MlNEnThrOverDouble. The
worst language (\MlEnThrErrHiLang) reaches \MlEnThrErrHi\%, while English stays within it (\MlEnThrErrEn\%). On XNLI
it misses the budget in \XnNEnThrOverTarget{} of \XnNLangs{} languages.

\paragraph{Own thresholds leave coverage unequal and often exceed the budget.} Under its own threshold, a MASSIVE
language stays within the budget on held-out items in \MlNOwnHold{} of \MlNLangs{} cases; among these, coverage
ranges from \MlOwnCovHoldLo\% to \MlOwnCovHoldHi\%, and the other \MlNOwnOverTarget{} exceed the budget
(Appendix~\ref{app:posthoc}). On XNLI an own threshold exists in \XnNOwnFeasible{} languages and exceeds the budget in
\XnNOwnOverTarget{} of them. Over \MlNRepSplits{} random half splits, the own threshold misses the budget in
\MlOwnOverRateMean\% of MASSIVE language--split pairs on average.

\paragraph{Neither procedure certifies a threshold on the exploratory split.} The sample is too small. Even a
threshold with no tuning errors must handle \MlLttNeedGrid{} tuning items to pass the Bonferroni test, against
\MlNTune{} per MASSIVE language, and the fixed-sequence variant needs \MlLttNeedOne{} handled items at $0.99$, where no
MASSIVE language handles more than \phMlFsNTopMax. The certified gate therefore defers every request, on XNLI as well
(Appendix~\ref{app:repro}).

\begin{table}[t]
\centering\small
\caption{Certification of per-language thresholds by label budget. Exploratory analysis of the confirmatory items,
\budNSplits{} random splits per language into $n$ tuning and at least \budNHeldMin{} held-out items. cert: share of
language--split pairs in which the fixed-sequence procedure certifies, with 90\% confidence, a threshold whose
population selective error is at most 5\%. cov en / mean: held-out coverage for English and the mean other language,
counting a split that certifies nothing as zero. own miss: share of pairs in which the uncertified own threshold
exceeds 5\% held-out selective error. In \%.}\label{tab:budget}
\setlength{\tabcolsep}{3pt}
\begin{tabular}{@{}r ccc c ccc c ccc@{}}
\toprule
 & \multicolumn{3}{c}{gpt-6-luna} && \multicolumn{3}{c}{\MsDec} && \multicolumn{3}{c}{\PxDec} \\
\cmidrule{2-4}\cmidrule{6-8}\cmidrule{10-12}
$n$ & cert & cov en / mean & own miss && cert & cov en / mean & own miss && cert & cov en / mean & own miss \\
\midrule
83 & \budLunaEightyThreeFsCert & \budLunaEightyThreeFsCovEn{} / \budLunaEightyThreeFsCovMean & \budLunaEightyThreeOwnMiss && \budMsEightyThreeFsCert & \budMsEightyThreeFsCovEn{} / \budMsEightyThreeFsCovMean & \budMsEightyThreeOwnMiss && \budPxEightyThreeFsCert & \budPxEightyThreeFsCovEn{} / \budPxEightyThreeFsCovMean & \budPxEightyThreeOwnMiss \\
150 & \budLunaOneFiftyFsCert & \budLunaOneFiftyFsCovEn{} / \budLunaOneFiftyFsCovMean & \budLunaOneFiftyOwnMiss && \budMsOneFiftyFsCert & \budMsOneFiftyFsCovEn{} / \budMsOneFiftyFsCovMean & \budMsOneFiftyOwnMiss && \budPxOneFiftyFsCert & \budPxOneFiftyFsCovEn{} / \budPxOneFiftyFsCovMean & \budPxOneFiftyOwnMiss \\
300 & \budLunaThreeHundredFsCert & \budLunaThreeHundredFsCovEn{} / \budLunaThreeHundredFsCovMean & \budLunaThreeHundredOwnMiss && \budMsThreeHundredFsCert & \budMsThreeHundredFsCovEn{} / \budMsThreeHundredFsCovMean & \budMsThreeHundredOwnMiss && \budPxThreeHundredFsCert & \budPxThreeHundredFsCovEn{} / \budPxThreeHundredFsCovMean & \budPxThreeHundredOwnMiss \\
500 & \budLunaFiveHundredFsCert & \budLunaFiveHundredFsCovEn{} / \budLunaFiveHundredFsCovMean & \budLunaFiveHundredOwnMiss && \budMsFiveHundredFsCert & \budMsFiveHundredFsCovEn{} / \budMsFiveHundredFsCovMean & \budMsFiveHundredOwnMiss && \budPxFiveHundredFsCert & \budPxFiveHundredFsCovEn{} / \budPxFiveHundredFsCovMean & \budPxFiveHundredOwnMiss \\
\bottomrule
\end{tabular}
\end{table}

\subsection{Thresholds certified with larger label budgets}
\paragraph{From 300 labels, fixed-sequence certification succeeds in more than half of language--split pairs on two
interfaces.} The larger confirmatory sample allows bigger label budgets (Table~\ref{tab:budget}; exploratory). The
fixed-sequence procedure certifies in \budMsThreeHundredFsCert\% and \budPxThreeHundredFsCert\% of language--split
pairs at 300 labels per language on \MsDec{} and \PxDec, against \budMsOneFiftyFsCert\% and \budPxOneFiftyFsCert\% at
150; gpt-6-luna reaches \budLunaFiveHundredFsCert\% only at 500. On a grid up to 0.99, and counting a split that
certifies nothing as zero coverage, English's held-out coverage is never below the mean other language's; at 500
labels \PxDec{} handles \budPxFiveHundredFsCovEn\% of English requests against \budPxFiveHundredFsCovMean\% for the mean
other language. Adding 1.00 to the grid reverses the comparison for gpt-6-luna at 500 labels
(Appendix~\ref{app:eqrisk}).

\subsection{Per-language conformal thresholds}
\paragraph{Per-language conformal thresholds bring set coverage to target but leave automatic handling unequal.} We
ran the conformal method of \citet{arxiv2609_37861_ax} on gpt-6-luna's outputs for the exploratory items. A request is handled automatically when its
prediction set holds a single label. With one threshold per language, set coverage reaches its \pxAlpha\% target in
every language on average, but English's lead over the lowest language in this singleton rate widens from
\pxMlSingPooledSpread{} points under a pooled threshold to \pxMlSingMondSpread{} on MASSIVE. Part of this widening follows from the two-decimal rounding of gpt-6-luna's
probabilities (Appendix~\ref{app:conformal}).

\section{Related work}

Work on multilingual calibration studies the probability values and how to repair them, for pretrained language
models and multilingual question answering models \citep{arxiv2210_12265,arxiv2311_08669_ax} and for LLMs
\citep{krause2023confidently,huang2026multilingualcalib}. \citet{feng2024abstain_tc} show that abstention strategies
developed on English lose accuracy in low-resource languages. We instead ask how the coverage of one fixed gate on a
production interface differs by language. Selective prediction and the reject option
\citep{chow1970reject,geifman2017selective} define coverage and selective risk, and \citet{jones2021selective} showed
that confidence-based abstention can widen group disparities. Our groups are the language versions of the same items
in production interfaces.

TypeSafe AI introduced the interface class with Jev \citep{typesafe2026jev}. Concurrently, a benchmark of Jev reports steep per-language accuracy degradation, with risk--coverage curves per dataset pooled
over languages \citep{arxiv2609_37647_ax}. Another benchmark compares the class with classifiers and language models
and applies held-out selective-risk thresholds \citep{arxiv2610_00346_ax}. Where \citet{arxiv2609_37647_ax} report
accuracy by language, we report what one fixed gate does in each language. \citet{arxiv2610_14112_ax} find on
Indonesian regional languages that Jev's calibration error does not grow relative to Indonesian while its ranking of
errors (area under the risk--coverage curve) worsens. We measure the coverage of one fixed threshold on three interfaces in \MlNLangs{} language
versions and test per-language thresholds, which they leave to future work. Also concurrently,
\citet{arxiv2609_37861_ax} show on MasakhaNEWS and AfriSenti that a pooled conformal threshold on a TF-IDF classifier
meets its 90\% coverage target on average but covers Somali at 77.5\%, and one threshold per language restores
coverage with 100--200 labels. \citet{arxiv2605_14260_ax} show that equal coverage and equal set size across groups are
in tension. Appendix~\ref{app:related} covers further related work.

\section{Limitations}\label{sec:lim}

The confirmatory study covers MASSIVE only. XNLI and the schema, format and open-weight checks
rest on \MlNPerLang{} or \XnNPerLang{} items per language, which resolves large coverage differences but not
individual ranks. Translate-test used one translator. MASSIVE has been public since 2022, so its English utterances may be better
represented in the models' training data than its translations.
In MASSIVE and XNLI every non-English version is a translation from English. The native-input check covers one task
and \injNHLangs{} African languages, with 31--40 human-translated items each, and measures the direction of the gap
for these languages. We measure deferral but not its cost: human-review availability, latency and outcomes are outside
the data.

\section{Conclusion}

On MASSIVE, a confidence gate on three typed decision interfaces serves 51 languages unequally: answers correct in
both languages are acted on more often in English, while the requests the gate handles are rarely wrong.
The gap remains after standardizing every language to a 75\% share of correct answers. In exploratory analyses it
also remains with a schema written in each input's language and with each language's threshold fitted to the same
5\% error target, and it holds for requests written natively in African languages, with English as the human
translation.
Translating inputs into English reduces the gap at $0.9$ from \cfPxDelta--\cfMsDelta{} to \ttDelta--\ttPxDelta{}
points; at a common 5\% error target, English's lead after translation is \eqttGapMin--\eqttGapMax{} points. At 300 labels per language, a fixed-sequence test certifies
per-language thresholds in more than half of language--split pairs on two interfaces, against under a third at 150. How often it certifies, and whether English keeps the higher coverage, depend on the
procedure and its threshold grid. We suggest that evaluations of typed decision interfaces
report, per language, coverage at the operating threshold and the mean paired handling gap against a reference
language, alongside calibration.

\section*{Broader Impact Statement}
Where selective error is controlled, a confidence gate can automate routine requests and reserve human review for
uncertain ones. A gate that defers more often in some languages gives their speakers less automation and, where
deferred requests go to human review, can mean longer waits. Per-language coverage reports can expose this before
deployment. The ``deferred with a correct top label'' counts are retrospective: a top label is known to be correct only
in hindsight. Lowering a language's threshold on their basis requires measuring the selective error at the new
threshold.

\bibliographystyle{tmlr}
\bibliography{P3}

\appendix
\renewcommand{\floatpagefraction}{0.8}
\FloatBarrier
\section{Confirmatory study details}
\label{app:confirm}
\paragraph{Plan.} Before the plan was written,
three requests per OpenRouter model on one already-used item (in English, Swahili and Javanese) checked the response
format. The plan fixes the following. The pool is the MASSIVE test ids present in all 51 languages minus every id used
in this paper's exploratory study or in an earlier 200-item pilot (Appendix~\ref{app:repro}), which leaves 2{,}633 eligible ids; the sample is \cfNItems{} ids drawn from the
pool sorted by id with NumPy's \texttt{default\_rng(20261015)}. Each request carries the MASSIVE intent question of
Section~\ref{sec:setup} alone, with the 59 intents of the test split as options. Scoring, the primary endpoint, its
bootstrap (\cfB{} resamples of the item ids with \texttt{default\_rng(20261016)}; a language whose $B_L$ is empty in a
resample is left out of that resample's mean), the two-sided bootstrap $p$-value $p = \min(1, 2\min(P^*(\Delta \le
0), P^*(\Delta \ge 0)))$, Holm's procedure over the three models and the decision rule are as in Section~\ref{sec:setup}. Requests
are retried on transport errors and HTTP 429 or 5xx, refusals are not, and all \cfNReq{} requests of a model are sent
with no interim look; an arm with fewer than 95\% answered requests counts as failed and enters no test. The secondary
measures, reported without a decision rule, are: per-language coverage at $0.9$ with the simultaneous band of its
difference from English; the same for ECE and for the share handled among correct answers; the coverage gap at a
common accuracy of $0.75$ (coverage recomputed as $0.75\,P(H \mid \text{correct}) + 0.25\,P(H \mid \text{wrong})$,
English minus the mean language, with a bootstrap interval); selective error per language with exact 95\% intervals;
the share of requests deferred with a correct top label; and, for gpt-6-luna, the Spearman correlation of
per-language coverage with the exploratory study. The ancillary files hold every secondary measure
for each model and language.

\paragraph{Models.} \MsDec{} and \PxDec{} were chosen from the OpenRouter models with
decision outputs, by four criteria: all 59 options returned with probabilities summing to one, full-precision
probabilities, a versioned model name and a vendor other than OpenAI. Checking the response format excluded models
that rejected a 59-option question, accepted at most 20 options or returned rounded probabilities. Each OpenRouter arm
then passed an admission check of 20 requests on exploratory items: every response returned all 59 options, with
probabilities summing to one within $0.01$ and a choice among the options. Requests go to
\url{https://openrouter.ai/api/v1/systemone} with the model name, the input, the question and
\texttt{"provider": \{"allow\_fallbacks": false\}}. gpt-6-luna's unanswered requests were refused or empty after retries.

\paragraph{Per-language results.} Table~\ref{tab:confirmlang} gives accuracy, coverage at $0.9$ and $d_L$ for every
language and model. The highest exact 95\% upper bound on selective error is \cfLunaSelErrUcbMax\% (gpt-6-luna, \cfLunaSelErrUcbMaxLang:
\cfLunaSelErrUcbMaxErr{} errors among \cfLunaSelErrUcbMaxHandled{} handled requests). On the \MlNPerLang{}
exploratory items (Table~\ref{tab:main}), gpt-6-luna's per-language coverage correlates with the confirmatory one at
Spearman \cfLunaRho.

\begin{table}[!htbp]
\centering\scriptsize\setlength{\tabcolsep}{4pt}
\caption{Confirmatory study per language. acc: accuracy; cov: coverage at $0.9$; $d_L$: among items correct in both English
and the language, the share handled in English minus the share handled in the language. All in \% or points.}\label{tab:confirmlang}
\begin{tabular}{@{}l ccc c ccc c ccc@{}}
\toprule
 & \multicolumn{3}{c}{gpt-6-luna} && \multicolumn{3}{c}{\MsDec} && \multicolumn{3}{c}{\PxDec} \\
\cmidrule{2-4}\cmidrule{6-8}\cmidrule{10-12}
lang & acc & cov & $d_L$ && acc & cov & $d_L$ && acc & cov & $d_L$ \\
\midrule
af & 72 & 42 & 18 &  & 82 & 63 & 11 &  & 80 & 64 & 10 \\
am & 70 & 43 & 17 &  & 74 & 52 & 18 &  & 74 & 50 & 23 \\
ar & 71 & 43 & 16 &  & 77 & 60 & 11 &  & 79 & 64 & 10 \\
az & 74 & 51 & 8 &  & 80 & 61 & 11 &  & 81 & 65 & 9 \\
bn & 74 & 48 & 12 &  & 80 & 60 & 14 &  & 83 & 67 & 9 \\
cy & 65 & 35 & 24 &  & 72 & 44 & 27 &  & 69 & 48 & 22 \\
da & 74 & 51 & 8 &  & 86 & 69 & 4 &  & 85 & 70 & 7 \\
de & 76 & 56 & 4 &  & 85 & 71 & 2 &  & 85 & 74 & 3 \\
el & 77 & 55 & 4 &  & 85 & 69 & 4 &  & 86 & 72 & 6 \\
en & 78 & 59 & -- &  & 88 & 75 & -- &  & 88 & 78 & -- \\
es & 75 & 53 & 6 &  & 84 & 70 & 4 &  & 84 & 72 & 5 \\
fa & 75 & 47 & 14 &  & 84 & 71 & 2 &  & 84 & 70 & 6 \\
fi & 73 & 46 & 15 &  & 83 & 63 & 11 &  & 84 & 69 & 7 \\
fr & 77 & 52 & 8 &  & 84 & 69 & 5 &  & 86 & 72 & 5 \\
he & 76 & 51 & 10 &  & 83 & 68 & 5 &  & 85 & 71 & 6 \\
hi & 77 & 53 & 7 &  & 83 & 66 & 8 &  & 84 & 72 & 5 \\
hu & 76 & 53 & 5 &  & 83 & 64 & 10 &  & 84 & 68 & 9 \\
hy & 74 & 49 & 11 &  & 83 & 63 & 11 &  & 81 & 65 & 10 \\
id & 76 & 52 & 8 &  & 84 & 69 & 5 &  & 85 & 73 & 4 \\
is & 72 & 43 & 17 &  & 76 & 55 & 17 &  & 78 & 62 & 10 \\
it & 76 & 53 & 7 &  & 86 & 72 & 3 &  & 86 & 74 & 4 \\
ja & 78 & 57 & 2 &  & 87 & 73 & 2 &  & 86 & 75 & 3 \\
jv & 64 & 27 & 37 &  & 70 & 47 & 23 &  & 73 & 55 & 18 \\
ka & 67 & 43 & 12 &  & 75 & 57 & 12 &  & 74 & 58 & 13 \\
km & 60 & 32 & 24 &  & 64 & 37 & 29 &  & 69 & 51 & 18 \\
kn & 74 & 48 & 11 &  & 78 & 56 & 14 &  & 81 & 66 & 9 \\
ko & 77 & 56 & 4 &  & 85 & 70 & 5 &  & 85 & 73 & 4 \\
lv & 74 & 50 & 11 &  & 82 & 61 & 12 &  & 80 & 66 & 8 \\
ml & 76 & 48 & 13 &  & 75 & 47 & 26 &  & 85 & 69 & 7 \\
mn & 76 & 50 & 10 &  & 74 & 54 & 20 &  & 78 & 61 & 15 \\
ms & 74 & 47 & 13 &  & 81 & 64 & 9 &  & 84 & 70 & 7 \\
my & 74 & 50 & 10 &  & 58 & 33 & 35 &  & 78 & 61 & 13 \\
nb & 75 & 49 & 11 &  & 84 & 66 & 8 &  & 84 & 68 & 10 \\
nl & 77 & 54 & 6 &  & 85 & 70 & 4 &  & 86 & 72 & 5 \\
pl & 78 & 55 & 5 &  & 85 & 71 & 2 &  & 87 & 74 & 4 \\
pt & 76 & 52 & 7 &  & 85 & 69 & 4 &  & 85 & 74 & 3 \\
ro & 75 & 47 & 14 &  & 84 & 65 & 8 &  & 83 & 69 & 7 \\
ru & 76 & 56 & 2 &  & 86 & 72 & 2 &  & 85 & 75 & 2 \\
sl & 73 & 48 & 12 &  & 84 & 61 & 13 &  & 82 & 67 & 8 \\
sq & 68 & 42 & 17 &  & 75 & 53 & 17 &  & 75 & 61 & 11 \\
sv & 76 & 52 & 8 &  & 84 & 68 & 5 &  & 86 & 72 & 5 \\
sw & 66 & 36 & 25 &  & 71 & 41 & 30 &  & 71 & 48 & 22 \\
ta & 74 & 46 & 13 &  & 65 & 33 & 39 &  & 82 & 66 & 11 \\
te & 77 & 49 & 9 &  & 78 & 54 & 19 &  & 83 & 68 & 9 \\
th & 74 & 48 & 12 &  & 76 & 56 & 17 &  & 81 & 68 & 8 \\
tl & 73 & 49 & 11 &  & 80 & 56 & 17 &  & 81 & 63 & 11 \\
tr & 77 & 54 & 4 &  & 85 & 67 & 7 &  & 85 & 72 & 4 \\
ur & 76 & 45 & 17 &  & 82 & 64 & 10 &  & 81 & 66 & 10 \\
vi & 76 & 54 & 7 &  & 85 & 71 & 3 &  & 86 & 73 & 4 \\
zh-CN & 77 & 54 & 6 &  & 84 & 71 & 4 &  & 84 & 74 & 3 \\
zh-TW & 74 & 48 & 10 &  & 81 & 67 & 5 &  & 82 & 70 & 5 \\
\bottomrule
\end{tabular}

\end{table}

\FloatBarrier
\section{Methods of the exploratory additions}
\label{app:newmethods}
\paragraph{Translate-test.} Each non-English confirmatory input was translated into English by \ttMTModel{} at
temperature 0, 50 utterances of one language per call, with the instruction ``Translate each of the following user
requests to English. Output only the translations, as a JSON object with the same keys.'' and the utterances as a JSON
object with numbered keys; the source language was not named. A reply was used only if it had exactly the requested
keys and a non-empty string for each; otherwise the call was retried, and after three failures each utterance was
translated in its own call with ``Translate the following user request to English. Output only the translation.''
Every interface received the translations with the unchanged English schema; English inputs and answers are those of
the confirmatory study. Similarity to the original English utterance is measured as exact match after lowercasing and
removing punctuation and extra whitespace (\ttRtExactShare\% of pairs; from \ttRtExactShareMin\% for
\ttRtExactShareMinLang{} to \ttRtExactShareMax\% for \ttRtExactShareMaxLang), and as character F-score (chrF;
mean \ttRtChrfMean). The low-similarity half is the pairs below the median chrF; gaps on both subsets use the
confirmatory bootstrap.

\begin{table}[!htbp]
\centering\small
\caption{The reduction from translate-test holds on fixed items and without translations that reproduce the English
utterance. Confirmatory items, gate at $0.9$. orig, TT: mean paired handling gap without and with translation;
change: TT minus orig with 95\% interval. Fixed items: items correct in both languages under both conditions. Least
similar half: pairs below the median chrF to the English utterance. In points.}\label{tab:ttcheck}
\setlength{\tabcolsep}{3pt}
\begin{tabular}{@{}lccc@{}}
\toprule
 & gpt-6-luna & \MsDec & \PxDec \\
\midrule
Fixed items, orig / TT & \ttIntDeltaOrig{} / \ttIntDelta & \ttMsIntDeltaOrig{} / \ttMsIntDelta & \ttPxIntDeltaOrig{} / \ttPxIntDelta \\
\quad change & \ttIntDiffDelta{} [\ttIntDiffDeltaLo, \ttIntDiffDeltaHi] & \ttMsIntDiffDelta{} [\ttMsIntDiffDeltaLo, \ttMsIntDiffDeltaHi] & \ttPxIntDiffDelta{} [\ttPxIntDiffDeltaLo, \ttPxIntDiffDeltaHi] \\
Without exact matches, orig / TT & \ttRtNonExactDeltaOrig{} / \ttRtNonExactDelta & \ttMsRtNonExactDeltaOrig{} / \ttMsRtNonExactDelta & \ttPxRtNonExactDeltaOrig{} / \ttPxRtNonExactDelta \\
\quad change & \ttRtNonExactDiffDelta{} [\ttRtNonExactDiffDeltaLo, \ttRtNonExactDiffDeltaHi] & \ttMsRtNonExactDiffDelta{} [\ttMsRtNonExactDiffDeltaLo, \ttMsRtNonExactDiffDeltaHi] & \ttPxRtNonExactDiffDelta{} [\ttPxRtNonExactDiffDeltaLo, \ttPxRtNonExactDiffDeltaHi] \\
Least similar half, orig / TT & \ttRtLowSimDeltaOrig{} / \ttRtLowSimDelta & \ttMsRtLowSimDeltaOrig{} / \ttMsRtLowSimDelta & \ttPxRtLowSimDeltaOrig{} / \ttPxRtLowSimDelta \\
\quad change & \ttRtLowSimDiffDelta{} [\ttRtLowSimDiffDeltaLo, \ttRtLowSimDiffDeltaHi] & \ttMsRtLowSimDiffDelta{} [\ttMsRtLowSimDiffDeltaLo, \ttMsRtLowSimDiffDeltaHi] & \ttPxRtLowSimDiffDelta{} [\ttPxRtLowSimDiffDeltaLo, \ttPxRtLowSimDiffDeltaHi] \\
\bottomrule
\end{tabular}
\end{table}

\paragraph{Natively written inputs.} The data are the test splits of InjongoIntent \citep{yu2025injongo}, published as
\texttt{masakhane/InjongoIntent} (revision \texttt{fe4be38}, Apache-2.0). Human pairs are the \injNH{} items of its
English test split, each a native utterance and its human English translation, with the language taken from the
example id (31--40 items per language). Machine pairs are the full test splits of Swahili, Amharic, Hausa, Yoruba and
Zulu, translated as in the translate-test arm above. Each request carries the MASSIVE intent question alone, with the
\injNIntents{} intents as options and each intent's name, underscores replaced by spaces, as its description. Each
interface receives \injNReq{} requests. gpt-6-luna refused \injLunaUnanswered{} of them,
scored as deferred and wrong; the other two interfaces answered every request. Intervals come from 10{,}000 resamples
of the items, within each language for the machine pairs.

\paragraph{Fine-tuned classifier.} The checkpoint is
\url{haryoaw/scenario-TCR-data-AmazonScience-massive-all_1.1-model-xlm-roberta-base} (revision
\texttt{f996499f}), an XLM-RoBERTa base model fine-tuned on the training split of MASSIVE in all its locales. Its 60
output classes map to MASSIVE's intent list; the one intent absent from the test split (\texttt{cooking\_query}) is
scored as wrong when predicted. Inputs are tokenized with the checkpoint's tokenizer and truncated to 128 tokens.
The confidence is the top softmax probability, and all other quantities follow the
confirmatory analysis with the same bootstrap draws. It handles \xlmHgcEn\% of correct English answers (the
interfaces: \cfLunaHgcEn--\cfPxHgcEn\%), its ECE is \xlmEceMin--\xlmEceMax\% in every language, and the simultaneous
band puts \xlmNCovBelow{} languages below English in coverage.

\paragraph{Label budgets.} For each budget $n$ of 83, 150, 300 and 500, \budNSplits{} random permutations of the
\cfNItems{} items are drawn once with a fixed seed and shared by all languages and interfaces; the first $n$ items of a permutation are
the tuning items and the rest are held out, so the budgets are nested. Thresholds are fitted as in
Section~\ref{sec:thr}. We use Learn-then-Test because it bounds selective error; conformal risk control \citep{angelopoulos2024crc} bounds
the share of all requests that are handled and wrong.

\FloatBarrier
\section{Full per-language tables}
\label{app:full}
Tables~\ref{tab:fullml} and~\ref{tab:fullxn} give every language, with the columns of Table~\ref{tab:main}.
\begin{table}[!htbp]
\centering\footnotesize\setlength{\tabcolsep}{3pt}
\caption{MASSIVE, all \MlNLangs{} languages, sorted by coverage under the English threshold. Columns as in
Table~\ref{tab:main}.}\label{tab:fullml}
\begin{tabular}{lrrrrrcc|lrrrrrcc}
\toprule
& & & & & paired & EN thr & own thr & & & & & & paired & EN thr & own thr \\
lang & acc & ECE & auto & err & hand/both & cov/err & cov/err & lang & acc & ECE & auto & err & hand/both & cov/err & cov/err \\
\midrule
tr & 80 & 5 & 58 & 1 & 6 / 18 & 76 / 2 & 84 / 5$\dagger$ & fi & 78 & 7 & 48 & 1 & 19 / 19 & 69 / 0 & 76 / 6$\dagger$ \\
vi & 80 & 5 & 53 & 0 & 14 / 17 & 76 / 2 & 78 / 2 & es & 76 & 4 & 47 & 3 & 22 / 17 & 69 / 2 & 33 / 0 \\
ko & 79 & 4 & 57 & 2 & 13 / 16 & 76 / 4 & 63 / 0 & th & 77 & 6 & 51 & 1 & 15 / 18 & 67 / 4 & 67 / 4 \\
de & 78 & 4 & 58 & 3 & 10 / 18 & 76 / 2 & 42 / 0 & tl & 76 & 4 & 48 & 1 & 17 / 18 & 67 / 4 & 70 / 4 \\
ru & 80 & 7 & 52 & 0 & 16 / 17 & 75 / 4 & 81 / 6$\dagger$ & zh-CN & 79 & 5 & 55 & 1 & 14 / 16 & 67 / 2 & 72 / 4 \\
hu & 77 & 5 & 51 & 3 & 16 / 19 & 75 / 6$\dagger$ & 76 / 8$\dagger$ & ta & 79 & 7 & 53 & 1 & 12 / 19 & 67 / 4 & 72 / 4 \\
hi & 78 & 5 & 54 & 1 & 11 / 18 & 75 / 6$\dagger$ & 78 / 6$\dagger$ & lv & 73 & 9 & 47 & 3 & 18 / 19 & 67 / 2 & 81 / 6$\dagger$ \\
pt & 79 & 5 & 51 & 0 & 16 / 18 & 73 / 4 & 81 / 7$\dagger$ & he & 78 & 8 & 46 & 3 & 20 / 21 & 67 / 2 & 66 / 2 \\
pl & 79 & 7 & 56 & 1 & 9 / 19 & 73 / 4 & 61 / 0 & af & 77 & 9 & 41 & 0 & 24 / 21 & 67 / 2 & 76 / 6$\dagger$ \\
nl & 83 & 8 & 53 & 0 & 14 / 19 & 73 / 2 & 87 / 5$\dagger$ & zh-TW & 76 & 9 & 49 & 5 & 18 / 20 & 66 / 7$\dagger$ & 31 / 0 \\
fr & 77 & 10 & 55 & 2 & 11 / 19 & 73 / 2 & 78 / 2 & kn & 75 & 8 & 45 & 1 & 22 / 17 & 64 / 5 & 66 / 5 \\
id & 77 & 5 & 55 & 4 & 12 / 16 & 72 / 4 & 54 / 0 & ur & 76 & 9 & 46 & 1 & 17 / 21 & 64 / 2 & 64 / 2 \\
bn & 77 & 5 & 51 & 3 & 17 / 17 & 72 / 4 & 79 / 9$\dagger$ & az & 74 & 6 & 49 & 0 & 14 / 20 & 63 / 2 & 70 / 4 \\
en & 83 & 5 & 60 & 1 & 0 / 28 & 72 / 0 & 72 / 0 & ms & 72 & 8 & 44 & 2 & 25 / 14 & 61 / 2 & 52 / 3 \\
fa & 78 & 10 & 47 & 1 & 19 / 21 & 72 / 2 & 79 / 4 & is & 75 & 6 & 43 & 0 & 19 / 20 & 61 / 0 & 61 / 0 \\
sv & 77 & 6 & 55 & 2 & 13 / 15 & 72 / 0 & 85 / 7$\dagger$ & el & 80 & 11 & 50 & 4 & 16 / 22 & 61 / 0 & 70 / 0 \\
ja & 83 & 6 & 54 & 1 & 16 / 15 & 72 / 0 & 73 / 0 & hy & 75 & 10 & 47 & 0 & 17 / 18 & 60 / 0 & 52 / 0 \\
te & 75 & 5 & 51 & 3 & 18 / 15 & 72 / 4 & 76 / 6$\dagger$ & ar & 74 & 6 & 44 & 2 & 20 / 20 & 60 / 2 & 69 / 4 \\
sl & 77 & 4 & 50 & 3 & 17 / 19 & 70 / 2 & 70 / 2 & am & 69 & 7 & 40 & 0 & 25 / 17 & 58 / 5$\dagger$ & 63 / 5 \\
mn & 77 & 8 & 51 & 3 & 16 / 16 & 70 / 6$\dagger$ & 75 / 6$\dagger$ & cy & 70 & 9 & 37 & 0 & 30 / 17 & 57 / 8$\dagger$ & 51 / 3 \\
ml & 78 & 6 & 49 & 4 & 21 / 15 & 70 / 6$\dagger$ & 75 / 6$\dagger$ & sq & 69 & 8 & 40 & 8 & 29 / 16 & 55 / 11$\dagger$ & 31 / 5 \\
da & 79 & 6 & 47 & 3 & 19 / 20 & 70 / 2 & 81 / 6$\dagger$ & ka & 67 & 7 & 45 & 4 & 16 / 18 & 55 / 5$\dagger$ & 52 / 3 \\
my & 73 & 8 & 50 & 1 & 15 / 17 & 69 / 2 & 76 / 4 & sw & 68 & 7 & 29 & 0 & 39 / 17 & 46 / 3 & 37 / 0 \\
ro & 75 & 7 & 45 & 0 & 23 / 16 & 69 / 4 & 69 / 4 & km & 65 & 4 & 32 & 4 & 30 / 21 & 46 / 6$\dagger$ & 36 / 4 \\
nb & 77 & 6 & 45 & 3 & 18 / 22 & 69 / 7$\dagger$ & 87 / 16$\dagger$ & jv & 64 & 11 & 24 & 3 & 49 / 13 & 42 / 7$\dagger$ & 45 / 7$\dagger$ \\
it & 82 & 5 & 55 & 0 & 10 / 19 & 69 / 0 & 79 / 4 & &&&&&&& \\
\bottomrule
\end{tabular}

\end{table}
\begin{table}[!htbp]
\centering\small
\caption{XNLI, all \XnNLangs{} languages. Columns as in Table~\ref{tab:main}, with \XnNTune{} tuning and \XnNTest{}
held-out items.}\label{tab:fullxn}
\begin{tabular}{lrrrrrcc}
\toprule
 & & & & & paired & EN thr & own thr \\
lang & acc & ECE & auto & err & hand/both & cov/err & cov/err \\
\midrule
en & 82 & 9 & 68 & 11 & 0 / 26 & 45 / 9$\dagger$ & 45 / 9$\dagger$ \\
ru & 69 & 20 & 61 & 23 & 18 / 12 & 44 / 22$\dagger$ & -- \\
bg & 76 & 12 & 61 & 15 & 16 / 13 & 40 / 11$\dagger$ & 31 / 9$\dagger$ \\
tr & 67 & 20 & 58 & 27 & 22 / 13 & 38 / 29$\dagger$ & -- \\
de & 76 & 13 & 61 & 17 & 13 / 16 & 38 / 13$\dagger$ & -- \\
vi & 67 & 20 & 58 & 21 & 17 / 11 & 38 / 19$\dagger$ & -- \\
th & 67 & 19 & 56 & 24 & 20 / 12 & 37 / 21$\dagger$ & -- \\
es & 75 & 12 & 57 & 16 & 17 / 15 & 37 / 13$\dagger$ & -- \\
zh & 68 & 20 & 60 & 24 & 17 / 12 & 37 / 17$\dagger$ & -- \\
el & 71 & 16 & 59 & 22 & 20 / 13 & 36 / 21$\dagger$ & 18 / 19$\dagger$ \\
fr & 72 & 15 & 60 & 14 & 15 / 13 & 35 / 20$\dagger$ & -- \\
hi & 66 & 19 & 55 & 28 & 26 / 10 & 33 / 21$\dagger$ & -- \\
sw & 65 & 21 & 56 & 28 & 24 / 9 & 30 / 23$\dagger$ & -- \\
ar & 65 & 20 & 56 & 26 & 19 / 15 & 29 / 17$\dagger$ & -- \\
ur & 64 & 19 & 47 & 29 & 32 / 13 & 26 / 24$\dagger$ & -- \\
\bottomrule
\end{tabular}

\end{table}
\FloatBarrier
\section{Schema translation checks}
\label{app:schema}
\begin{figure}[!htbp]
\centering
\includegraphics[width=0.48\linewidth]{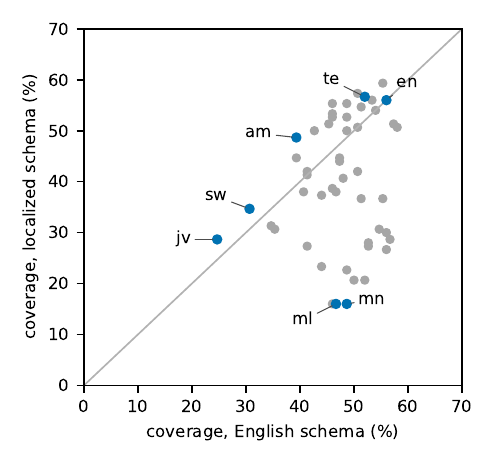}
\caption{For gpt-6-luna, localizing the schema lowers coverage in more languages than it raises it. Each point is
one language: coverage at $0.9$ with the English ($x$) and the localized schema ($y$).}
\label{fig:loc}
\end{figure}

\paragraph{Localized schema.} The schemas are LLM translations, checked by back-translation and, for Swahili, by
LLM review (below). The baseline is an
English-schema run of the same items from the same session (run-to-run variation in Appendix~\ref{app:repro}). On
gpt-6-luna the two coverage rankings barely correlate (Spearman \LocSpearmanAuto), mean coverage changes by
\LocMeanDAuto{} points, and English's lead over the lowest language grows from \LocSpreadEn{} to \LocSpreadLoc{}
points (Figure~\ref{fig:loc}). For the other two interfaces the ranking barely moves (Spearman \lorMsSpearmanAuto{}
and \lorPxSpearmanAuto), and English's lead stays similar: \lorMsSpreadEn{} to \lorMsSpreadLoc{} points for \MsDec{}
and \lorPxSpreadEn{} to \lorPxSpreadLoc{} for \PxDec.

\begin{table}[!htbp]
\centering\small
\caption{In most of these languages, localizing the schema changes coverage more than accuracy or selective error. Same items with the English
schema (EN) and with the schema in the input's language (local), in \%: accuracy of the intent question, coverage
at $0.9$ (auto@0.9) and selective error at $0.9$.}\label{tab:loc}
\begin{tabular}{lrrrrrr}
\toprule
 & \multicolumn{2}{c}{accuracy} & \multicolumn{2}{c}{auto@0.9} & \multicolumn{2}{c}{sel.\ error} \\
\cmidrule(lr){2-3}\cmidrule(lr){4-5}\cmidrule(lr){6-7}
lang & EN & local & EN & local & EN & local \\
\midrule
en & 83 & 83 & 56 & 56 & 1 & 1 \\
nl & 82 & 81 & 51 & 51 & 0 & 0 \\
de & 78 & 76 & 57 & 51 & 2 & 3 \\
zh-CN & 78 & 79 & 53 & 27 & 0 & 0 \\
ja & 84 & 84 & 56 & 27 & 0 & 0 \\
hi & 77 & 79 & 53 & 56 & 1 & 4 \\
ar & 73 & 74 & 44 & 37 & 2 & 0 \\
ur & 77 & 74 & 46 & 55 & 1 & 2 \\
am & 69 & 74 & 39 & 49 & 0 & 1 \\
jv & 67 & 67 & 25 & 29 & 3 & 2 \\
sw & 67 & 71 & 31 & 35 & 0 & 0 \\
cy & 70 & 71 & 35 & 31 & 2 & 2 \\
\bottomrule
\end{tabular}

\end{table}
Table~\ref{tab:loc} gives selected languages under both schemas. Its English-schema baseline is a run sent in the same
session as the localized one; English's lead over the lowest language there is \LocSpreadEn{} points, against
\rcvMlGateSpread{} in the first run of Table~\ref{tab:main}, a run-to-run difference. With the localized schema, coverage ranges over
\LocAutoLo--\LocAutoHi\%. Swahili is handled automatically \LocAutoSw\% of the time against \LocAutoEn\% for
English, and the English--Swahili difference goes from \LocGapEnSwEn{} to \LocGapEnSwLoc{} points. The ten languages
lowest under the English schema change by \LocMeanDAutoBottom{} points on average and the other languages by
\LocMeanDAutoRest, so the new bottom is made of languages that fell, not of the old bottom left behind. For gpt-6-luna, coverage rises by more than two points in \LocNAutoUp{} languages and falls by more than two in
\LocNAutoDown, and selective error under the localized schema stays at \LocErrLo--\LocErrHi\%.
Under the localized schemas, accuracy moved by at most \LocDAccAbsMax{} points in any language
and by \LocDAccMn{} for \LocMaxLossLang, the language that lost the most coverage. The accuracy change barely
correlates with the coverage change (Pearson \LocCorrDAccDAuto{} across the non-English languages).

\paragraph{Mean paired handling gap.} Each run's gap uses its own English answers, with 95\% intervals from \cfB{}
resamples of the items. With the English and the localized schema, the gap is \lorLuDeltaEn{} [\lorLuDeltaEnLo,
\lorLuDeltaEnHi] and \lorLuDeltaLoc{} [\lorLuDeltaLocLo, \lorLuDeltaLocHi] points for gpt-6-luna, \lorMsDeltaEn{}
[\lorMsDeltaEnLo, \lorMsDeltaEnHi] and \lorMsDeltaLoc{} [\lorMsDeltaLocLo, \lorMsDeltaLocHi] for \MsDec, and
\lorPxDeltaEn{} [\lorPxDeltaEnLo, \lorPxDeltaEnHi] and \lorPxDeltaLoc{} [\lorPxDeltaLocLo, \lorPxDeltaLocHi] for
\PxDec. The 95\% intervals of the change are [\lorLuDeltaDiffLo, \lorLuDeltaDiffHi],
[\lorMsDeltaDiffLo, \lorMsDeltaDiffHi] and [\lorPxDeltaDiffLo, \lorPxDeltaDiffHi] points.

\paragraph{Back-translation.} An automatic back-translation check with a different LLM (\pbtModel{}
\citep{openai2026gpt54mini}) over all \pbtNItems{} schema strings per language finds a mean equivalence of
\pbtJudgeMean{} of 5 on a rubric from 5 (same meaning) to 1 (unrelated or wrong), with \pbtShareLeTwo\% of strings
judged changed (score 2 or less; \pbtNInstrLow{} of them instructions). These are spread over \pbtNLangsLowItem{} of the \pbtNLangs{} languages, and the ten languages with the most hold only
\pbtShareLowTopTen\% of them. A language's judge drift is 5 minus its mean score and its lexical drift is 1 minus the mean token F1 between
each English string and its back-translation. We computed \pbtNTests{} rank correlations between these drift measures and the per-language changes in accuracy and coverage, with nominal $p$-values. Neither measure has a detected association with the coverage change under localization: Spearman
\pbtRhoJudgeAuto{} ($p=\pbtPJudgeAuto$) for judge drift and \pbtRhoLexAuto{} ($p=\pbtPLexAuto$) for lexical drift
(label descriptions alone \pbtRhoLexAutoLabels, $p=\pbtPLexAutoLabels$, Bonferroni \pbtPLexAutoLabelsBonf). Mongolian (\LocMaxLossLang), the largest coverage loss, ranks \pbtDriftRankMn{} of \pbtNLangs{} on judge drift, tied with two other languages.

\paragraph{English rewording.} The rewording restates the question text and each of the 59 intent descriptions in
different English words with the same meaning; option values and items are unchanged. We say that localization
reorders the languages more than the rewording when the 95\% interval of the paired difference between the two
Spearman correlations lies above zero. The reworded, localized and
English-schema runs were sent in one session; the endpoint refused the intent question in
\pphNRefused{} reworded requests. Accuracy changed by \pphMeanDAccPara{} points on average, and selective error at
$0.9$ stays at or below \pphErrHiPara\% in every language. The rewording's Spearman correlation with the original
ordering is \pphRhoPara; the paired difference between it and the localized schema's correlation is \pphRhoDiff{}
(95\% interval [\pphRhoDiffLo, \pphRhoDiffHi]), and the rewording raises mean coverage by \pphMeanDPara{} points.
The reworded schema's gate decisions agree with the baseline run's on \pphGateAgreePara\% of requests, and the
localized schema's on \pphGateAgreeLoc\%; two sendings of the unchanged suite on different days agree on
\pcoRerunGateAgree\% (Appendix~\ref{app:repro}).

\paragraph{Swahili review and corrected-schema run.} Two automated LLM reviews of the Swahili schema rated 9 and 8 of its \swfixNEntries{} strings wrong and 20 and 19
imprecise. The \swfixNEntries{} strings are the 60 that the single-question request uses plus the scenario question, 18
scenario names and the Boolean template of the multi-question request. All three alarm intents, for instance, used
\emph{alama} (mark) for alarm. The corrected schema takes the \swfixNFromReviews{} strings that both reviews
proposed identically and a fresh \pbtModel{} translation of the other \swfixNFromAzure; the corrected-schema run sends
its question text and intent descriptions with the unchanged option values. On the same \swfixN{} Swahili items, coverage went from \swfixAutoOrig\% to \swfixAutoFix\% (paired
difference \swfixDAutoOrig{} points, 95\% bootstrap interval [\swfixDAutoOrigLo, \swfixDAutoOrigHi]). Accuracy went
from \swfixAccOrig\% to \swfixAccFix\% (interval of the difference [\swfixDAccOrigLo, \swfixDAccOrigHi]). ECE went
from \swfixEceOrig\% to \swfixEceFix\% (jointly answered items: [\swfixDEceOrigLo, \swfixDEceOrigHi]). The
English--Swahili coverage gap went from \swfixGapOrig{} to \swfixGapFix{} points (\swfixGapEn{} under the English
schema).
\section{Reproducibility details}
\label{app:repro}

\begin{sloppypar}
\paragraph{Requests.} Each gpt-6-luna request is one \texttt{POST} to \url{https://api.openai.com/v1/decisions}
with the JSON body \texttt{\{"model": "gpt-6-luna", "input": <input>, "questions": [...]\}}. No other field is sent,
so no decoding setting is chosen; the endpoint returns probabilities rounded to two decimals and reports zero output
tokens. Identical requests are not fully deterministic, as reported for repeated runs of LLMs under deterministic
settings \citep{ouyang2025nondeterminism,atil2025nondeterminism}. The exploratory MASSIVE suite, sent a second time
on another day, keeps the top label in \pcoRerunTopAgree\% and the gate decision at $0.9$ in \pcoRerunGateAgree\% of
the \pcoRerunN{} requests answered both times. Within one session the variation is smaller: the localized run's
\pcoLocEnN{} English requests, identical to those of the English-schema run sent in the same session, returned the same
intent probabilities in \pcoLocEnSame{} cases. The input is the MASSIVE utterance as plain text, or for XNLI the object
\texttt{\{"premise": ..., "hypothesis": ...\}} serialized as indented JSON, since the endpoint takes text only. A Choice
question is sent as \texttt{\{"type": "choice", "name", "instructions", "choices": [\{"value", "description"\}]\}}, a
Boolean as \texttt{\{"type": "predicate", "name", "instructions"\}}. Every response names the model
\texttt{gpt-6-luna}. All requests were sent in October 2026, with 8 concurrent requests and a 120\,s timeout. The OpenRouter requests of the confirmatory study are described in
Appendix~\ref{app:confirm}; their probabilities are not rounded.
\end{sloppypar}

\paragraph{Questions.} MASSIVE intent (Choice): ``What does the user want the assistant to do?'', one option per
intent label of the test split (59), with the label as value and the label with underscores replaced by spaces as
description (\texttt{alarm\_set}: ``alarm set''). Scenario (Choice, multi-question format only): ``Which area does the user's request belong
to?'', the 18 label prefixes as value and description. Boolean, sent only in the multi-question format: ``Is the user asking the assistant to do this:
$\langle$target description$\rangle$?''. XNLI (Choice): ``How does the hypothesis relate to the premise?'' with
entailment ``the hypothesis definitely follows from the premise'', neutral ``the hypothesis might or might not be
true given the premise'' and contradiction ``the hypothesis definitely contradicts the premise''. Boolean: ``Does
the hypothesis relate to the premise in this way: $\langle$description$\rangle$?''. The localized schemas translate
these strings and keep the option values.

\paragraph{Items and Boolean targets.} The confirmatory study's items are drawn as in Appendix~\ref{app:confirm}. The
exploratory MASSIVE items come from the test split of all \MlNLangs{} languages
(\texttt{mteb/amazon\_massive\_intent}): the ids present in every language, in the order of the English test split, are shuffled by Python's
\texttt{random} with seed 20260919 and the first \MlNPerLang{} kept. The pilot excluded from the confirmatory pool is
built the same way from the ids present in 11 languages (en, zh-CN, zh-TW, ja, ko, es, de, ar, hi, vi, sw), shuffled
with seed 20260917, first 200 kept. XNLI items (\texttt{facebook/xnli}) are the first \XnNPerLang{} of the shuffle of the test indices $0, 1, \ldots$
under the same seed, with a fresh generator. For each item the Boolean target is drawn once, with the same generator: the gold
label with probability 0.5, otherwise a uniformly drawn other label, and the same target is used in every language. The Boolean question appears only in the multi-question format.

\paragraph{Available material.} The ancillary files contain the aggregate values of every printed table and the
generated values behind the results reported in the text: aggregate statistics per system and language, and the per-item values of the worked example in Figure~\ref{fig:example}. The items come
from the public MASSIVE, XNLI and InjongoIntent test splits and are selected by the procedure above and in
Appendices~\ref{app:confirm} and~\ref{app:newmethods}; the open-weight models are the public repositories
named below, run at each repository's default revision at run time.

\paragraph{Failures and refusals.} A request gets up to 8 attempts on a connection error or HTTP 429, 529 or
$\ge 500$, waiting $\min(60, 2^a)$ seconds times a uniform factor in $[0.5, 1.5]$ after attempt $a$; any other
non-200 status is stored as a failure. No exploratory request failed. The endpoint refused the intent
question in \MlNRefused{} of \MlNRequests{} MASSIVE requests (\LocNRefused{} localized) and the XNLI question in
\XnNRefused{} of \XnNRequests; refusals are scored as described in Section~\ref{sec:setup} (Refusals).

\paragraph{Open-weight models.} Each model received the multi-question request and ran through the inference code in its Hugging Face repository
(\texttt{Mapika/decider-2b}, \texttt{Hanno-Labs/bosun-v3.1-1.7b}, \texttt{FrontiersMind/Lumma-fev-0.6b}), which
takes the same input and questions and returns a probability per option, so prompt formatting and probability
extraction are the repositories' own. \texttt{open-decider} uses the repository's \texttt{Decider.system\_one},
\texttt{open-lumma} its \texttt{decide}, and \texttt{open-bosun} its \texttt{predict}, called once per question with
the options as candidates (it permutes their order internally). Settings are the repositories' defaults, the XNLI
input is passed as the structured pair, and the gate reads the top option probability as for the API. The
comparator of Section~\ref{sec:rob} is \texttt{convaiinnovations/laya-multilingual}, read the same way.

\paragraph{Statistics.} ECE is top-label: confidences fall into 10 equal-width bins on $[0, 1]$, each closed on the
left (a confidence of 0.7 opens the bin $[0.7, 0.8)$), and ECE is the sum
over non-empty bins of the bin's share of items times $|$mean confidence $-$ accuracy$|$, refusals excluded. With
\MlNPerLang{} exploratory items per language the bins hold 15 items on average and typically one holds fewer than
five, so per-language ECE there is noisy and biased \citep{roelofs2022mitigating}; the confirmatory study's \cfNItems{} items
give 100 per bin on average. The confirmatory study's simultaneous bands use the construction below on the \cfB{} resamples of
its primary bootstrap.

For the exploratory analyses, paired
rates rest on the items correct in both English and the language, \MlNBothRightHandoffHi{} for \MlHandoffHiLang. Marginal and paired 95\% intervals are percentile intervals from 2{,}000 resamples of the items (paired:
of the items shared with English). The simultaneous band uses \phSimB{} draws, each one multiset of items
shared by all languages. The fixed-threshold coverage comparisons with English and the comparison of handled
correct answers use one run of these draws; the risk--coverage bands come from one run of
\rcvNBoot{} draws, and the single- vs multi-question format-change band from \phNBoot{} draws. Let $d_k$ be the observed difference of language $k$ against English, $d^*_k$ its value in a
draw and $s_k$ the standard deviation of $d^*_k$ over draws. Then $q$ is the 95\% quantile of
$\max_k |d^*_k - d_k| / s_k$, and the band is $d_k \pm q s_k$ ($q = \MlEceDiffSimQ$ for ECE and \MlAutoDiffSimQ{}
for coverage on MASSIVE, \XnEceDiffSimQ{} and \XnAutoDiffSimQ{} on XNLI). In the exploratory study a language's refused pairs drop out of its
ECE contrast; in the confirmatory study each language's ECE is computed on its own answered requests. The per-threshold bands use the same construction at each threshold. The rank-correlation
intervals of Appendix~\ref{app:posthoc} resample languages and items jointly (\phNBoot{} draws); those of the
English rewording resample items only, with the \MlNLangs{} languages fixed. The corrected-schema run uses 4{,}000 paired
resamples. The repeated threshold splits of the 150-item exploratory study use 200 random halves drawn separately for each language, with the English
threshold fitted on the same items as the language's tuning half. The conformal comparison uses 200 splits shared
by all languages. Random generators are seeded per system and language.

\paragraph{Threshold fits.} The empirical fit is the lowest observed confidence whose selective error on the tuning
items is at most 5\% (integer counts). The Learn-then-Test fit tests each threshold $t$ of the grid $0.50, 0.51,
\dots, 0.99$ by $p_t = P(\mathrm{Bin}(n_t, 0.05) \le e_t)$, with $n_t$ tuning items at or above $t$ and $e_t$ errors
among them, and certifies the lowest $t$ with $p_t \le 0.1/50$ (Bonferroni over the grid). The fixed-sequence variant tests from 0.99 downward at level 0.1 and stops at the first failure. It certifies a
threshold in \phMlFsNCertified{} of \phMlFsNLangs{} MASSIVE and \phXnFsNCertified{} of \phXnFsNLangs{} XNLI languages: 0.99
already fails everywhere on the exploratory even/odd split, and at 0.99 no MASSIVE language has more than \phMlFsNTopMax{} tuning items (\MlLttNeedOne{}
error-free items are needed).

\FloatBarrier
\section{Threshold sweep and further checks}
\label{app:posthoc}
\paragraph{Marginal and unadjusted intervals at $0.9$.} These descriptive comparisons complement the simultaneous
band of Section~\ref{sec:res}. On MASSIVE every non-English marginal ECE interval overlaps English's. The marginal
coverage intervals of English [\MlAutoEnLo, \MlAutoEnHi]\% and Swahili [\MlAutoLoSw, \MlAutoHiSw]\% do not overlap,
and those of \MlNAutoBelowEn{} languages lie entirely below English's interval. Unadjusted paired intervals exclude
zero for \MlNEceDiffResolved{} (ECE, with upper ends of up to \MlEceDiffHiMax{} points) and \MlNAutoDiffResolved{}
(coverage) languages. On XNLI unadjusted intervals detect an ECE difference for \XnNEceDiffResolved{} languages,
against \XnNEceDiffResolvedSim{} under the simultaneous band. Across the \phXnNLangs{} XNLI languages, ECE and
coverage rank-correlate at \phXnRhoEceCov{} (95\% interval [\phXnRhoEceCovLo, \phXnRhoEceCovHi], resampling languages
and items). The interval spans both signs, so the direction of the association is unresolved.

\begin{figure}[!htbp]
\centering
\includegraphics[width=\linewidth]{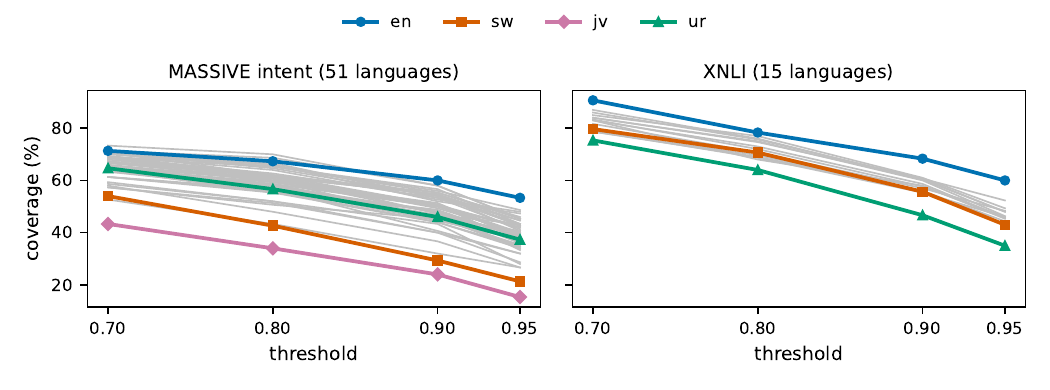}
\caption{The language ordering changes little across thresholds. Coverage on MASSIVE and
XNLI at the four tested thresholds $0.7$, $0.8$, $0.9$ and $0.95$, one gray line per language, with English, Swahili, Javanese and
Urdu highlighted.}
\label{fig:sweep}
\end{figure}

\paragraph{Threshold sweep.} The coverage gap is present at every tested threshold (Figure~\ref{fig:sweep}).
On MASSIVE, coverage ranges over \MlAutoSeventyLo--\MlAutoSeventyHi\% at $0.7$ and \MlAutoNinetyFiveLo--\MlAutoNinetyFiveHi\%
at $0.95$. At $0.7$, $0.8$, $0.9$ and $0.95$ the mean English-minus-language gap is \phSeventyMeanGap{} (95\%
interval [\phSeventyMeanGapLo, \phSeventyMeanGapHi]), \phEightyMeanGap{} [\phEightyMeanGapLo, \phEightyMeanGapHi],
\phNinetyMeanGap{} [\phNinetyMeanGapLo, \phNinetyMeanGapHi] and \phNinetyFiveMeanGap{} [\phNinetyFiveMeanGapLo,
\phNinetyFiveMeanGapHi] points. The English--Swahili gap is \phSeventySwGap{} [\phSeventySwLo, \phSeventySwHi],
\phEightySwGap{} [\phEightySwLo, \phEightySwHi], \phNinetySwGap{} [\phNinetySwLo, \phNinetySwHi] and
\phNinetyFiveSwGap{} [\phNinetyFiveSwLo, \phNinetyFiveSwHi] points, each interval from that threshold's own
simultaneous band. Javanese is the lowest at every threshold on MASSIVE.

\paragraph{Resource classes.} The English-minus-language gap at $0.9$ rank-correlates with the
resource class of \citet{joshi-etal-2020-state}, a rating of how much data and tooling a language has (higher is
better resourced), assigned by language name on MASSIVE. The rank correlation is \phJoshiRho, with a 95\% interval of [\phJoshiRhoLo, \phJoshiRhoHi] that
resamples languages and items as if the \phJoshiN{} languages were a sample. Better-resourced languages tend to trail
English by less, but the relation is not monotonic. By class, from 1 to 5 (5 the best resourced), the
mean English-minus-language gap at $0.9$ is \phJoshiGapB, \phJoshiGapC, \phJoshiGapD, \phJoshiGapE{} and
\phJoshiGapF{} points over \phJoshiNB, \phJoshiNC, \phJoshiND, \phJoshiNE{} and \phJoshiNF{} languages. The
\phJoshiNC{} class-2 languages trail English by more than the class-1 languages, and class~5 by more than
class~4. Slovenian, listed in two classes, takes class 3, and Norwegian (\texttt{nb}) takes the Norwegian Bokm{\aa}l class, 1.

\paragraph{Own-threshold misses.} Table~\ref{tab:ownfit} lists every language whose own threshold, fitted on the
tuning items, has held-out selective error above 5\%, with the handled and error counts behind it. The intervals
are per language, without simultaneous correction across languages. Only \rcvMlOwnWorstLang's exact interval among MASSIVE languages, and \rcvXnOwnWorstLang's on XNLI, lie wholly
above 5\% (Table~\ref{tab:ownfit}). The English threshold fitted
on the tuning half, \phMlEnFitThr, differs from the in-sample \rcvMlFiveThrEn{} of Section~\ref{sec:res} because it
uses the tuning half only.

\begin{table}[!htbp]
\centering\footnotesize\setlength{\tabcolsep}{4pt}
\caption{Most own-threshold misses are not distinguishable from 5\%. Languages whose own threshold (fitted on the
\MlNTune{} MASSIVE or \XnNTune{} XNLI tuning items to a 5\% budget) has held-out selective error above 5\%. thr: the
threshold; $n$: the held-out requests it handles; $e$: the errors among them; err: the error rate, with its exact
(Clopper--Pearson) 95\% interval (CI), in \%. Held-out items: \rcvMlOwnNTest{} (MASSIVE) and \rcvXnOwnNTest{}
(XNLI) per language.}\label{tab:ownfit}
\begin{tabular}{llrrrrl|llrrrrl}
\toprule
suite & lang & thr & $n$ & $e$ & err & 95\% CI & suite & lang & thr & $n$ & $e$ & err & 95\% CI \\
\midrule
MASSIVE & nl & 0.51 & 58 & 3 & 5.2 & [1.1, 14.4] & MASSIVE & mn & 0.63 & 50 & 3 & 6.0 & [1.3, 16.5] \\
MASSIVE & tr & 0.44 & 56 & 3 & 5.4 & [1.1, 14.9] & MASSIVE & jv & 0.72 & 30 & 2 & 6.7 & [0.8, 22.1] \\
MASSIVE & lv & 0.58 & 54 & 3 & 5.6 & [1.2, 15.4] & MASSIVE & sv & 0.45 & 57 & 4 & 7.0 & [1.9, 17.0] \\
MASSIVE & da & 0.61 & 54 & 3 & 5.6 & [1.2, 15.4] & MASSIVE & pt & 0.51 & 54 & 4 & 7.4 & [2.1, 17.9] \\
MASSIVE & ru & 0.55 & 54 & 3 & 5.6 & [1.2, 15.4] & MASSIVE & hu & 0.71 & 51 & 4 & 7.8 & [2.2, 18.9] \\
MASSIVE & hi & 0.62 & 52 & 3 & 5.8 & [1.2, 15.9] & MASSIVE & bn & 0.54 & 53 & 5 & 9.4 & [3.1, 20.7] \\
MASSIVE & af & 0.48 & 51 & 3 & 5.9 & [1.2, 16.2] & MASSIVE & nb & 0.42 & 58 & 9 & 15.5 & [7.3, 27.4] \\
MASSIVE & fi & 0.66 & 51 & 3 & 5.9 & [1.2, 16.2] & XNLI & bg & 0.99 & 45 & 4 & 8.9 & [2.5, 21.2] \\
MASSIVE & te & 0.65 & 51 & 3 & 5.9 & [1.2, 16.2] & XNLI & en & 0.98 & 65 & 6 & 9.2 & [3.5, 19.0] \\
MASSIVE & ml & 0.67 & 50 & 3 & 6.0 & [1.3, 16.5] & XNLI & el & 1.00 & 26 & 5 & 19.2 & [6.6, 39.4] \\
\bottomrule
\end{tabular}

\end{table}

\paragraph{Smaller label budgets and repeated splits.} The own fit is repeated on a random subset of 25 or 50 tuning
items. With 25 tuning items the fit exceeds twice the budget in \MlNBudgetTFOverDouble{} MASSIVE languages, and with
50 items in \MlNBudgetFOverDouble. We also repeat the fit over \MlNRepSplits{} random \MlNRepHalf/\MlNRepHalf{}
splits of the same \MlNPerLang{} items, which overlap. The own
threshold misses the budget on the held-out half in \MlOwnOverRateMean\% of language--split pairs on average
(\MlOwnOverRateLo--\MlOwnOverRateHi\% per language, English \MlOwnOverRateEn\%), and in more than half the splits for
\MlNOwnOverMajority{} of \MlNLangs{} languages. Counting splits in which no threshold meets the budget on the tuning
half as misses also gives \MlOwnOverOrInfeasibleMean\%. The English threshold misses in \MlEnOverRateMean\% of pairs.

\paragraph{Uncorrected Learn-then-Test bounds.} Without correcting for the choice of threshold, the tightest 90\%
bound that any single threshold reaches on the tuning items is close to the budget at best: \MlLttMinUcbLo\%
(\MlLttMinUcbLoLang) to \MlLttMinUcbHi\% (\MlLttMinUcbHiLang) on MASSIVE, English \MlLttMinUcbEn\%, and
\XnLttMinUcbLo--\XnLttMinUcbHi\% on XNLI.

\subsection{Multi-question requests}
\label{app:choiceonly}
The multi-question request adds two questions to the gate's Choice question: on MASSIVE a scenario question (``Which
area does the user's request belong to?'', 18 options) and a Boolean question asking whether the user wants one given
intent, and on XNLI the Boolean question only (Appendix~\ref{app:repro}). The gate reads only the answer to its own
question. The open-weight models of Section~\ref{sec:rob} received this format. gpt-6-luna received it for the same
\pcoMlCmpNPairs{} MASSIVE and \pcoXnCmpNPairs{} XNLI requests, and the analysis is the same. Per-language coverage at $0.9$ correlates across the two formats at
Spearman \pcoMlCmpRhoCov{} on MASSIVE and \pcoXnCmpRhoCov{} on XNLI, and the English-minus-language gaps at
\pcoMlCmpRhoGap{} and \pcoXnCmpRhoGap. The mean coverage change is \pcoMlCmpDCovMean{} points on MASSIVE (95\%
interval [\pcoMlCmpDCovMeanLo, \pcoMlCmpDCovMeanHi]) and \pcoXnCmpDCovMean{} on XNLI ([\pcoXnCmpDCovMeanLo,
\pcoXnCmpDCovMeanHi]). Per-language changes range from \pcoMlCmpDCovMin{} to \pcoMlCmpDCovMax{} points on MASSIVE,
and the simultaneous band detects \pcoMlCmpNDCovResolvedSim{} of them (XNLI: \pcoXnCmpNDCovResolvedSim). The answers
differ between the formats about as often as two sendings of the same multi-question request do
(Table~\ref{tab:choiceonly}). In both formats English has the highest coverage at $0.9$ and Javanese the lowest, no
language is detected above English, the simultaneous band detects no MASSIVE ECE difference, and no Learn-then-Test threshold is
certified.

\begin{table}[!htbp]
\centering\small
\caption{Single-question (main) and multi-question requests on the same items. Brackets are 95\% bootstrap intervals;
``simult.'' counts languages whose paired difference against English the simultaneous 95\% band detects.
Top-label agreement and distance count the \pcoMlCmpNPairsAnswered{} MASSIVE and \pcoXnCmpNPairsAnswered{} XNLI
requests answered in both formats; gate agreement counts all requests, with a refusal deferred. The last column compares
the single-question answer and a re-sent multi-question copy with the original multi-question answer on the same
\pcoRepN{} MASSIVE requests.}\label{tab:choiceonly}
\setlength{\tabcolsep}{3pt}\footnotesize\begin{tabular}{@{}lcc@{}}
\toprule
MASSIVE unless noted & single (main) & multi \\
\midrule
Coverage at $0.9$, English / Swahili / Javanese (\%) & \pcoMlAutoEnCo{} / \pcoMlAutoSwCo{} / \pcoMlAutoJvCo{} & \pcoMlAutoEnMq{} / \pcoMlAutoSwMq{} / \pcoMlAutoJvMq{} \\
Languages below 50\% coverage at $0.9$ & \pcoMlNAutoBelowHalfCo{} & \pcoMlNAutoBelowHalfMq{} \\
Below en at $0.7$/$0.8$/$0.9$/$0.95$ (simult.) & \pcoMlSeventyNBelowSimCo{} / \pcoMlEightyNBelowSimCo{} / \pcoMlNinetyNBelowSimCo{} / \pcoMlNinetyFiveNBelowSimCo{} & \pcoMlSeventyNBelowSimMq{} / \pcoMlEightyNBelowSimMq{} / \pcoMlNinetyNBelowSimMq{} / \pcoMlNinetyFiveNBelowSimMq{} \\
Above English at the same thresholds & \pcoMlSeventyNAboveSimCo{} / \pcoMlEightyNAboveSimCo{} / \pcoMlNinetyNAboveSimCo{} / \pcoMlNinetyFiveNAboveSimCo{} & \pcoMlSeventyNAboveSimMq{} / \pcoMlEightyNAboveSimMq{} / \pcoMlNinetyNAboveSimMq{} / \pcoMlNinetyFiveNAboveSimMq{} \\
Mean English-minus-language gap at $0.9$ (points) & \pcoMlNinetyMeanGapCo{} [\pcoMlNinetyMeanGapLoCo, \pcoMlNinetyMeanGapHiCo] & \pcoMlNinetyMeanGapMq{} [\pcoMlNinetyMeanGapLoMq, \pcoMlNinetyMeanGapHiMq] \\
ECE differences detected (simult. / unadjusted) & \pcoMlNEceDiffResolvedSimCo{} / \pcoMlNEceDiffResolvedCo{} & \pcoMlNEceDiffResolvedSimMq{} / \pcoMlNEceDiffResolvedMq{} \\
Own threshold within 5\% held out (of \pcoMlNLangsCo) & \pcoMlNOwnHoldCo{} & \pcoMlNOwnHoldMq{} \\
Learn-then-Test thresholds certified & \pcoMlLttNFeasibleCo{} & \pcoMlLttNFeasibleMq{} \\
XNLI coverage at $0.9$, English / Urdu (\%) & \pcoXnAutoEnCo{} / \pcoXnAutoUrCo{} & \pcoXnAutoEnMq{} / \pcoXnAutoUrMq{} \\
XNLI ECE above English (simult., of \pcoXnNDiffLangsCo) & \pcoXnNEceDiffAboveSimCo{} & \pcoXnNEceDiffAboveSimMq{} \\
\midrule
Agreement with multi-question run & single question, MASSIVE / XNLI & re-sent: single / copy \\
\midrule
Top label agrees (\%) & \pcoMlCmpTopAgree{} / \pcoXnCmpTopAgree{} & \pcoRepCoTopAgree{} / \pcoRepTopAgree{} \\
Gate decision at $0.9$ agrees (\%) & \pcoMlCmpGateAgree{} / \pcoXnCmpGateAgree{} & \pcoRepCoGateAgree{} / \pcoRepGateAgree{} \\
Mean total-variation distance per item & \pcoMlCmpTvMean{} / \pcoXnCmpTvMean{} & \pcoRepCoTvMean{} / \pcoRepTvMean{} \\
\bottomrule
\end{tabular}
\end{table}

\FloatBarrier
\section{Risk--coverage details}
\label{app:riskcov}
\paragraph{Exploratory items.} With every MASSIVE language at its own in-sample 5\% threshold, English reaches
\rcvMlFiveEn\% coverage and leads the mean language by \rcvMlFiveMeanGap{} points (95\% interval [\rcvMlFiveMeanGapLo,
\rcvMlFiveMeanGapHi]), although the simultaneous band detects no single language below English. Because those
thresholds are chosen in-sample, we also fix the coverage instead. On XNLI, where no language's full-sample curve
reaches 5\% observed error, selective error at 50\% coverage is \rcvXnHalfEn\% for English and up to \rcvXnHalfHi\%
(\rcvXnHalfHiLang), and the band puts \rcvXnHalfNAboveSim{} of \pcoXnNDiffLangsCo{} languages above English
(Table~\ref{tab:riskcov}).

For each language, the requests are sorted by the gate's confidence and the threshold is lowered one distinct value
at a time, so tied requests enter together. Coverage at a risk level is the largest coverage whose observed selective
error is at most that level (zero if no threshold reaches it), and the threshold reported is the one that attains it.
Selective error at a common coverage takes the most confident share of requests, with a tied group entering pro rata
(the expectation under random tie-breaking). Intervals come from
\rcvNBoot{} resamples of the items, each shared by all languages, and the simultaneous bands use the construction of
Appendix~\ref{app:repro} on the differences from English. Table~\ref{tab:riskcov} lists every language.

On XNLI, \rcvXnTenNReach{} of \XnNLangs{} languages reach 10\% observed selective error at some threshold (English at
\rcvXnTenEn\% coverage), and none reaches 5\%. At 50\% coverage on MASSIVE, selective error is \rcvMlHalfEn\% for English and
up to \rcvMlHalfHi\% (\rcvMlHalfHiLang); the mean language exceeds English by \rcvMlHalfMeanGap{} points
([\rcvMlHalfMeanGapLo, \rcvMlHalfMeanGapHi]), and the simultaneous band detects no single language. Among the
open-weight models, \texttt{open-decider} reaches 5\% observed selective error at \rcvMlDeciderFiveEn\% coverage in
English and \rcvMlDeciderFiveLo--\rcvMlDeciderFiveHi\% across languages, and the simultaneous band puts
\rcvMlDeciderFiveNBelowSim{} languages below English. For \texttt{open-bosun} the values are \rcvMlBosunFiveEn\% and
\rcvMlBosunFiveLo--\rcvMlBosunFiveHi\%, with \rcvMlBosunFiveNBelowSim{} languages detected.

\begin{table}[!htbp]
\centering\scriptsize
\caption{Risk--coverage per language, gpt-6-luna. auto, err: coverage and selective error (errors/handled) at
$0.9$. cov at risk: the largest coverage whose observed selective error is at most 5\% (MASSIVE, top) or 10\%
(XNLI, bottom), with its 95\% bootstrap interval; thr: the threshold that attains it (``--'': none). XNLI also gives
selective error at a common coverage of 50\% (err at 50\% cov) with its 95\% bootstrap interval, and its difference
from English in points with the simultaneous 95\% band ($-$ en [simult.]). 0 [0, 0]: no
threshold meets the budget in the sample or in any resample. In \%.}\label{tab:riskcov}
\setlength{\tabcolsep}{3pt}\begin{tabular}{lrrrr|lrrrr}
\toprule
lang & auto & err & cov at risk $\le 5$\% & thr & lang & auto & err & cov at risk $\le 5$\% & thr \\
\midrule
sq & 40 & 5/60 & 31 [4, 59] & 0.94 & fa & 47 & 1/71 & 70 [59, 80] & 0.57 \\
km & 32 & 2/48 & 33 [13, 60] & 0.89 & ml & 49 & 3/74 & 71 [7, 82] & 0.67 \\
jv & 24 & 1/36 & 42 [8, 57] & 0.72 & mn & 51 & 2/77 & 71 [33, 81] & 0.60 \\
sw & 29 & 0/44 & 44 [31, 67] & 0.79 & bn & 51 & 2/76 & 71 [34, 81] & 0.56 \\
zh-TW & 49 & 4/73 & 47 [24, 82] & 0.91 & id & 55 & 3/83 & 71 [43, 83] & 0.65 \\
cy & 37 & 0/55 & 51 [39, 65] & 0.77 & lv & 47 & 2/70 & 71 [45, 80] & 0.58 \\
ka & 45 & 3/68 & 51 [15, 71] & 0.77 & my & 50 & 1/75 & 71 [60, 80] & 0.49 \\
nb & 45 & 2/67 & 57 [39, 82] & 0.84 & af & 41 & 0/61 & 72 [54, 84] & 0.49 \\
am & 40 & 0/60 & 59 [43, 73] & 0.59 & zh-CN & 55 & 1/82 & 72 [61, 83] & 0.62 \\
ms & 44 & 1/66 & 59 [41, 79] & 0.74 & da & 47 & 2/71 & 72 [41, 83] & 0.61 \\
is & 43 & 0/65 & 61 [47, 75] & 0.67 & hi & 54 & 1/81 & 73 [53, 82] & 0.62 \\
ro & 45 & 0/67 & 62 [47, 74] & 0.74 & hy & 47 & 0/70 & 73 [49, 83] & 0.55 \\
es & 47 & 2/70 & 63 [9, 77] & 0.75 & pt & 51 & 0/76 & 73 [63, 85] & 0.63 \\
tl & 48 & 1/72 & 63 [45, 75] & 0.70 & el & 50 & 3/75 & 73 [41, 87] & 0.56 \\
kn & 45 & 1/68 & 63 [50, 83] & 0.73 & sv & 55 & 2/82 & 73 [57, 84] & 0.56 \\
hu & 51 & 2/76 & 67 [43, 81] & 0.73 & ta & 53 & 1/80 & 73 [58, 85] & 0.60 \\
ko & 57 & 2/85 & 67 [16, 85] & 0.76 & de & 58 & 3/87 & 73 [35, 85] & 0.70 \\
he & 46 & 2/69 & 67 [33, 82] & 0.65 & fr & 55 & 2/82 & 74 [36, 82] & 0.57 \\
th & 51 & 1/76 & 67 [53, 79] & 0.68 & vi & 53 & 0/80 & 74 [62, 84] & 0.57 \\
fi & 48 & 1/72 & 67 [56, 78] & 0.66 & it & 55 & 0/83 & 76 [60, 87] & 0.56 \\
sl & 50 & 2/75 & 68 [24, 81] & 0.65 & ja & 54 & 1/81 & 77 [64, 88] & 0.62 \\
ar & 44 & 1/66 & 68 [53, 76] & 0.50 & ru & 52 & 0/78 & 77 [62, 88] & 0.53 \\
az & 49 & 0/73 & 69 [53, 78] & 0.54 & nl & 53 & 0/79 & 78 [59, 89] & 0.53 \\
ur & 46 & 1/69 & 69 [53, 81] & 0.60 & en & 60 & 1/90 & 80 [67, 90] & 0.52 \\
te & 51 & 2/76 & 69 [32, 81] & 0.65 & tr & 58 & 1/87 & 80 [57, 87] & 0.44 \\
pl & 56 & 1/84 & 69 [56, 85] & 0.72 &  & & & &  \\
\bottomrule
\end{tabular}

\medskip
\begin{tabular}{lrrrrrr}
\toprule
lang & auto & err & cov at risk $\le 10$\% & thr & err at 50\% cov & $-$ en [simult.] \\
\midrule
ar & 56 & 43/168 & 0 [0, 27] & -- & 23 [16, 30] & 15.1 [5.8, 24.4] \\
de & 61 & 32/183 & 0 [0, 46] & -- & 17 [11, 23] & 9.8 [1.5, 18.0] \\
el & 59 & 38/176 & 0 [0, 22] & -- & 19 [13, 27] & 11.8 [4.0, 19.6] \\
es & 57 & 27/171 & 0 [0, 53] & -- & 15 [10, 21] & 7.9 [0.4, 15.3] \\
fr & 60 & 26/180 & 0 [0, 60] & -- & 15 [10, 21] & 7.6 [0.7, 14.6] \\
hi & 55 & 46/166 & 0 [0, 18] & -- & 27 [20, 34] & 19.7 [9.9, 29.4] \\
ru & 61 & 42/182 & 0 [0, 0] & -- & 21 [15, 29] & 13.9 [4.9, 22.8] \\
sw & 56 & 47/167 & 0 [0, 13] & -- & 27 [20, 34] & 19.3 [9.0, 29.5] \\
th & 56 & 40/168 & 0 [0, 25] & -- & 22 [15, 29] & 14.0 [5.5, 22.6] \\
tr & 58 & 47/173 & 0 [0, 19] & -- & 26 [19, 33] & 18.5 [9.8, 27.1] \\
ur & 47 & 40/140 & 0 [0, 0] & -- & 29 [22, 37] & 21.6 [11.1, 32.0] \\
vi & 58 & 36/174 & 0 [0, 28] & -- & 18 [13, 25] & 10.8 [2.3, 19.2] \\
zh & 60 & 43/181 & 0 [0, 19] & -- & 21 [15, 29] & 13.9 [4.9, 22.9] \\
bg & 61 & 27/182 & 53 [0, 65] & 0.93 & 8 [4, 14] & 1.0 [\ensuremath{-}4.9, 6.8] \\
en & 68 & 23/205 & 60 [0, 81] & 0.95 & 8 [4, 12] & -- \\
\bottomrule
\end{tabular}

\end{table}

\FloatBarrier
\section{Thresholds fitted to a common risk target, and a certification grid up to 1.00}
\label{app:eqrisk}
Both analyses are exploratory and use the confirmatory items. For coverage at a common risk target, a language's threshold is the
lowest confidence whose selective error is at most 5\%, as in Appendix~\ref{app:riskcov}. The in-sample threshold is
fitted on all \cfNItems{} items. The held-out threshold is the own threshold of Section~\ref{sec:thr}, fitted on
\eqNHeld{} tuning items of each of the \budNSplits{} splits of the label-budget analysis and applied to the other
\eqNHeld; held-out coverage is the mean over splits. Intervals and simultaneous bands use the confirmatory study's
item resamples: all \cfB{} in-sample and the first \eqBHeld{} held-out, with every threshold refitted in each resample
(Table~\ref{tab:eqrisk}). With held-out thresholds the simultaneous band puts \eqHeldNBelowMin--\eqHeldNBelowMax{}
languages below English, and held-out selective error exceeds 5\% in \eqMissMin--\eqMissMax\% of language--split
pairs, scattering around the target.

The certification comparison applies the fixed-sequence procedure of Table~\ref{tab:budget} with two grids: 0.50 to
0.99, and 0.50 to 1.00, where testing starts at 1.00. A threshold of 1.00 handles only requests with probability 1.00. gpt-6-luna returns
probabilities rounded to two decimals, so these exist; \MsDec{} and \PxDec{} return none, and their test fails at the
first step (Table~\ref{tab:certtop}).

\paragraph{Label budgets in detail.} The Bonferroni variant certifies far less than the fixed-sequence one: at 500
labels in \budMsFiveHundredLttCert\% of pairs for \MsDec, \budPxFiveHundredLttCert\% for \PxDec{} and
\budLunaFiveHundredLttCert\% for gpt-6-luna. On the grid up to 0.99, English is certified more often than the average
other language: at 500 labels, \PxDec{} is certified for English in \budPxFiveHundredFsCertEn\% of splits but for
\budPxFiveHundredFsCertLowLang{} in \budPxFiveHundredFsCertLow\%. Held-out coverage of the certified gate falls both
when certification fails and when the certified threshold is high; at 500 labels gpt-6-luna handles
\budLunaFiveHundredFsCovEn\% of English requests and \budLunaFiveHundredFsCovMean\% of the mean other language's,
against \cfLunaCovEn\% of English requests under the fixed $0.9$ gate. The uncertified own threshold exceeds 5\%
held-out selective error in \budLunaFiveHundredOwnMiss--\budLunaEightyThreeOwnMiss\% of gpt-6-luna's pairs, depending
on the label budget, and in a similar share for the other two interfaces. A grid that also includes 1.00 certifies
nothing for \MsDec{} and \PxDec, whose probabilities stay below 1.00, and certifies gpt-6-luna at 500 labels in
\eqLunaFiveHundredTopCert\% of pairs instead of \eqLunaFiveHundredOrigCert\%; English's coverage
(\eqLunaFiveHundredTopCovEn\%) then falls below the mean other language's (\eqLunaFiveHundredTopCovMean\%)
(Table~\ref{tab:certtop}).

\begin{table}[!htbp]
\centering\scriptsize
\caption{A fixed-sequence grid that also includes 1.00 certifies less often than the grid up to 0.99 wherever the latter certifies. Confirmatory
items, splits as in Table~\ref{tab:budget}. cert: share of language--split pairs certified, grid to 0.99 / grid to 1.00. cov: held-out
coverage for English / the mean other language under each grid, counting a split that certifies nothing as zero.
In \%.}\label{tab:certtop}
\setlength{\tabcolsep}{3pt}\begin{tabular}{@{}r ccc c ccc c ccc@{}}
\toprule
 & \multicolumn{3}{c}{gpt-6-luna} && \multicolumn{3}{c}{\MsDec} && \multicolumn{3}{c}{\PxDec} \\
\cmidrule{2-4}\cmidrule{6-8}\cmidrule{10-12}
$n$ & cert & cov, 0.99 & cov, 1.00 && cert & cov, 0.99 & cov, 1.00 && cert & cov, 0.99 & cov, 1.00 \\
\midrule
83 & 0 / 0 & 0 / 0 & 0 / 0 &  & 0.2 / 0 & 0.6 / 0.1 & 0 / 0 &  & 4 / 0 & 23 / 2 & 0 / 0 \\
150 & 2 / 0 & 11 / 0.8 & 0 / 0 &  & 30 / 0 & 29 / 15 & 0 / 0 &  & 27 / 0 & 46 / 15 & 0 / 0 \\
300 & 11 / 3 & 14 / 4 & 3 / 0.8 &  & 63 / 0 & 61 / 32 & 0 / 0 &  & 55 / 0 & 71 / 31 & 0 / 0 \\
500 & 18 / 10 & 24 / 6 & 2 / 3 &  & 87 / 0 & 74 / 45 & 0 / 0 &  & 78 / 0 & 78 / 46 & 0 / 0 \\
\bottomrule
\end{tabular}

\end{table}

\FloatBarrier
\section{Conformal comparison details}
\label{app:conformal}
\paragraph{Method.} We ran the method of \citet{arxiv2609_37861_ax} on our outputs (Table~\ref{tab:conformal}).
Split conformal prediction with the LAC score (one minus the probability of the true label) returns for each request
a set of labels meant to contain the true one with \pxAlpha\% probability; this probability is the set coverage. A
request is handled automatically when its set holds a single label, and the share of such requests is the singleton
rate, the conformal counterpart of our coverage. The threshold is calibrated on all languages pooled or on each
language, over \pxNReps{} random half splits. Because the pooled rows are translations of the same items, the pooled
threshold has no finite-sample guarantee.

\paragraph{Results.} Their finding recurs on our outputs. The pooled threshold falls below target set coverage in \pxMlNCovPooledBelow{} of
\pxMlNLangs{} MASSIVE languages (lowest \pxMlCovPooledRepLo\%, \pxMlCovPooledRepLoLang) and in \pxXnNCovPooledBelow{}
of \pxXnNLangs{} XNLI languages, while per-language thresholds bring every language to \pxMlCovMondRepMin\% or more
on average (XNLI \pxXnCovMondRepMin\%). The singleton rate stays unequal: English's lead over the lowest language
widens from \pxMlSingPooledSpread{} points under the pooled threshold to \pxMlSingMondSpread{} under per-language
thresholds on MASSIVE, against \pxMlGateSpread{} for our gate on the same items (XNLI: \pxXnSingPooledSpread{},
\pxXnSingMondSpread{} and \pxXnGateSpread). These leads are averaged over splits, so they differ slightly from the
differences of the averaged rates in Table~\ref{tab:conformal}. Set coverage also flags different languages than our
gate: of the \pxMlBotCovPooledSize{} languages lowest in pooled set coverage, \pxMlBotCovPooledOverlap{} are among the
\pxMlBotGateSize{} lowest at our gate (ties included).

Per-language conformal thresholds therefore trade unequal set coverage for a wider gap in singleton rates, which
parallels the tension \citet{arxiv2605_14260_ax} derive between equal coverage and equal set size. For gpt-6-luna part
of this widening is mechanical: its probabilities are rounded to two decimals, and on the even/odd split
\pxMlNQOwnMax{} of the \pxMlNLangs{} per-language thresholds admit every intent with probability at least
\pxMlPKeepMin, so a singleton occurs only when all other intents round to zero. The APS score shows the same pattern
with a smaller widening.

\begin{table}[!htbp]
\centering\small
\caption{Per-language conformal thresholds bring set coverage to target in every language, on average over splits, but
leave singleton rates unequal. gpt-6-luna, exploratory items. The method of \citet{arxiv2609_37861_ax} on MASSIVE (LAC score, \pxAlpha\% target),
calibrated on all languages pooled or on each language, next to our gate at $0.9$. Averaged over \pxNReps{} random half
splits shared by all languages. cov: held-out set coverage. single: share of held-out items whose set holds one label.
err (conformal): error among singletons. auto, err: the gate's coverage and selective error. correct, deferred: share
deferred although the top label was correct, under per-language conformal sets (conf.) and under the gate. The gate
columns are held-out averages over the same splits, so they differ slightly from Table~\ref{tab:main}.
$\dagger$: set coverage below \pxAlpha\% before rounding. In \%.}\label{tab:conformal}
\begin{tabular}{lrrrrrrrrrr}
\toprule
 & \multicolumn{3}{c}{pooled conformal} & \multicolumn{3}{c}{per-language conformal} & \multicolumn{2}{c}{gate $0.9$} & \multicolumn{2}{c}{correct, deferred} \\
\cmidrule(lr){2-4}\cmidrule(lr){5-7}\cmidrule(lr){8-9}\cmidrule(lr){10-11}
lang & cov & single & err & cov & single & err & auto & err & conf. & gate \\
\midrule
en & 93 & 53 & 1 & 91 & 56 & 2 & 60 & 1 & 28 & 23 \\
sw & 87$\dagger$ & 27 & 2 & 95 & 7 & 0 & 30 & 0 & 61 & 39 \\
jv & 86$\dagger$ & 22 & 3 & 96 & 5 & 0 & 24 & 3 & 60 & 41 \\
km & 84$\dagger$ & 30 & 5 & 94 & 8 & 0 & 32 & 4 & 57 & 34 \\
cy & 89$\dagger$ & 33 & 3 & 93 & 18 & 0 & 37 & 0 & 52 & 33 \\
am & 89$\dagger$ & 35 & 0 & 93 & 20 & 0 & 40 & 0 & 49 & 29 \\
sq & 89$\dagger$ & 33 & 7 & 93 & 19 & 4 & 40 & 8 & 52 & 33 \\
\bottomrule
\end{tabular}

\end{table}

\paragraph{Splits and results.} Each of \pxNReps{} draws splits the items into halves, and the same split applies to
every language, so a held-out item never enters calibration through one of its translations. One threshold is
calibrated on the calibration half of all languages pooled (\pxMlNCal{} items in \pxMlNLangs{} languages,
\pxMlNCalPooled{} rows on MASSIVE) or one per language (\pxMlNCal{} items). With $n$ calibration scores the
threshold is the $\lceil (n+1) \times \pxAlpha\% \rceil$-th smallest, and a set holds every label whose score is at most the
threshold. The pooled threshold covers the true
intent in \pxMlCovPooledRepMean\% of held-out items on average over languages. Under per-language thresholds set
coverage is \pxMlCovMondRepLo--\pxMlCovMondRepHi\% in every language, while the singleton rate runs from
\pxMlSingMondLo\% (\pxMlSingMondLoLang) to \pxMlSingMondHi\% (\pxMlSingMondHiLang), English \pxMlSingMondEn\%. Under
the same thresholds \pxMlDeferCorrectMondSw\% of held-out Swahili requests are deferred although the top label is
correct, against \pxMlDeferCorrectMondEn\% for English (gate: \pxMlGateDeferCorrectSw\% and
\pxMlGateDeferCorrectEn\%). A refusal gets score 1, so its set is empty below threshold 1 and holds every label at 1. Under per-language
thresholds, mean set size reaches \pxMlSizeMondHi{} intents (English \pxMlSizeMondEn). On the even/odd split it is up to
\pxMlSizeMondParHi{} (English \pxMlSizeMondParEn). Mean singleton error under per-language thresholds is at most
\pxMlSingErrMondHi\% in any MASSIVE language and reaches \pxXnSingErrMondHi\% on XNLI (\pxXnSingErrMondHiLang), since
\pxAlpha\% set coverage does not hold the error of the singletons alone to \pxMiss\%.

\paragraph{APS score.} The sensitivity check replaces LAC with the randomized APS score
\citep{romano2020aps}: the labels are ranked by probability, with tied labels in a random order, and the score is
the probability of the labels ranked above the true one plus a uniform random fraction of the true label's own
probability. The order and the fraction are drawn once per request. Splits, calibration halves and the singleton rule are
those of the LAC comparison. Under APS the pooled threshold falls short of \pxAlpha\% set coverage in
\rcvApsMlNCovPooledBelow{} of \pxMlNLangs{} MASSIVE languages (lowest \rcvApsMlCovPooledMin\%) and in
\rcvApsXnNCovPooledBelow{} of \pxXnNLangs{} XNLI languages. The English-minus-lowest singleton gap is
\rcvApsMlSingPooledSpread{} points pooled and \rcvApsMlSingMondSpread{} per language on MASSIVE (lowest per language:
\rcvApsMlSingMondLoLang, \rcvApsMlSingMondLo\%), and \rcvApsXnSingPooledSpread{} and \rcvApsXnSingMondSpread{} on XNLI.
Per-language calibration thus widens the singleton gap under both scores, by less under APS (LAC: \pxMlSingPooledSpread{}
to \pxMlSingMondSpread{} on MASSIVE).

\paragraph{Localized schema.} With the localized schema, \pxLocBotSingMondOverlap{} of the
\pxMlBotSingMondSize{} languages with the lowest per-language singleton rate stay among the \pxLocBotSingMondSize{}
lowest (ties included), and English leads the lowest language by \pxLocSingMondSpread{} points
(\pxMlSingMondSpread{} with the English schema).

\FloatBarrier
\section{Open-weight models and the small comparator}
\label{app:open}
\begin{table}[!htbp]
\centering\small
\caption{English ranks first or second in coverage in every model, and its observed selective error is no greater than
that of the other languages pooled. MASSIVE exploratory items at $0.9$, in \%: coverage (auto) and selective error
(err, $k$ errors among $n$ handled) for English, the lowest-coverage language (tied languages pooled) and all non-English languages pooled; spread:
en $-$ lowest; $\rho$: Spearman correlation of per-language coverage with gpt-6-luna's. The open-weight rows use the
multi-question request; the gpt-6-luna row and $\rho$ use its single-question run.}\label{tab:mlcross}
\setlength{\tabcolsep}{4pt}\begin{tabular}{lrrrrrrr}
\toprule
 & \multicolumn{2}{c}{en} & \multicolumn{2}{c}{lowest language} & other languages & & \\
\cmidrule(lr){2-3}\cmidrule(lr){4-5}
system & auto & err (k/n) & auto (lang) & err (k/n) & err (k/n) & en $-$ lowest & $\rho$ \\
\midrule
\texttt{gpt-6-luna} & 60 & 1 (1/90) & 24 (jv) & 3 (1/36) & 2 (68/3{,}601) & 36 & 1.00 \\
\texttt{open-bosun} & 58 & 7 (6/87) & 1 (am) & 100 (2/2) & 14 (396/2{,}862) & 57 & 0.76 \\
\texttt{open-decider} & 45 & 0 (0/67) & 1 (km) & 0 (0/1) & 2 (30/1{,}537) & 44 & 0.54 \\
\texttt{open-lumma} & 13 & 5 (1/20) & 0 (12 tied) & -- (0/0) & 22 (52/236) & 13 & 0.47 \\
\bottomrule
\end{tabular}

\end{table}

\paragraph{Coverage ranking.} Three open-weight
models of the same interface class, \texttt{open-decider} \citep{mapika2026decider}, \texttt{open-bosun}
\citep{hanno2026bosun} and \texttt{open-lumma} \citep{frontiersmind2026lumma}, received the same items in the
multi-question format (Table~\ref{tab:mlcross}). English's rank by coverage at $0.9$ is \rkBosunEnRank{} for \texttt{open-bosun} (behind \rkBosunAboveEn),
\rkDeciderEnRank{} for \texttt{open-decider}, \rkLummaEnRank{} for \texttt{open-lumma} (tied with \ifnum\rkLummaEnTies=1
one other language\else\rkLummaEnTies{} other languages\fi) and \rkSvcEnRank{} for gpt-6-luna. For
\texttt{open-decider} and gpt-6-luna, English's higher coverage comes at similar observed selective error to the
other languages. For \texttt{open-bosun} and \texttt{open-lumma}, selective error is also higher outside English
(\rcvXBosunErrOther\% and \rcvXLummaErrOther\% pooled, against \rcvXBosunErrEn\% and \rcvXLummaErrEn\%). These
models handle few requests in their lowest languages (Table~\ref{tab:mlcross}), so their rankings are coarse.

English's lead over the median language at $0.9$ is \rkBosunEnOverMedian, \rkDeciderEnOverMedian{} and
\rkLummaEnOverMedian{} points for \texttt{open-bosun}, \texttt{open-decider} and \texttt{open-lumma}, and
\rkSvcEnOverMedian{} for gpt-6-luna. A language's marginal 95\% coverage interval lies entirely below English's
interval in \MlXBosunNBelowEn, \MlXDeciderNBelowEn{} and \MlXLummaNBelowEn{} languages for the three open models, and
in \MlXSvcNBelowEn{} for gpt-6-luna (its single-question run, as in Table~\ref{tab:mlcross}). This is a descriptive count without pairing or multiplicity adjustment. The
language orderings correlate with gpt-6-luna's at Spearman \MlXBosunRho, \MlXDeciderRho{} and \MlXLummaRho. Of each
model's ten lowest languages, ties included, \MlXBosunBottomOverlap, \MlXDeciderBottomOverlap{} and
\MlXLummaBottomOverlap{} are among gpt-6-luna's ten lowest. The \texttt{open-lumma} model has \MlXLummaBottomSize{}
languages at zero coverage, so its bottom set is not ten, and these languages cannot be ordered among themselves.

The three models differ in overall accuracy and confidence. The \texttt{open-decider} model handles \MlXDeciderEn\% of
English automatically, and its English accuracy of \MlXDeciderAccEn\% is above gpt-6-luna's. The \texttt{open-lumma}
model handles \MlXLummaEn\% (English accuracy \MlXLummaAccEn\%). On XNLI only one of the three clears the threshold
often enough to compare languages. The \texttt{open-decider} model handles \MlXXDeciderEn\% of English (gpt-6-luna
\MlXXSvcEn\%), and its XNLI ordering has a Spearman correlation of \MlXXDeciderRho{} with gpt-6-luna's. The
\texttt{open-bosun} model clears the threshold on about \MlXXBosunMax\% of requests at best, and \texttt{open-lumma}
on a single item in one language.

\paragraph{Small comparator.} laya-multilingual \citep{convai2026layaml} is a small open multilingual model of the
interface class (about 0.3B parameters, Apache-licensed), run on the same items. On MASSIVE it is a weak classifier
(intent accuracy \LyMlAccLo--\LyMlAccHi\%, English \LyMlAccEn\%), and its coverage tracks its accuracy
(\LyMlAutoLo--\LyMlAutoHi\%). On XNLI it is overconfident everywhere: it handles \LyXnAutoLo--\LyXnAutoHi\% of
requests automatically in every language, with selective error from \LyXnAutoErrEn\% (English) to
\LyXnAutoErrHi\% and ECE up to \LyXnEceHi\%. On MASSIVE, gpt-6-luna's gate shows the opposite pattern: selective
error stays within \MlAutoErrLo--\MlAutoErrHi\% while coverage varies by language.

\FloatBarrier
\section{Further related work}
\label{app:related}
\citet{huang2026multilingualcalib} find that instruction tuning on high-resource-language data raises confidence in
low-resource languages without a matching gain in accuracy, and \citet{alfarano2026uncertainty} compare uncertainty
estimation methods for multiple-choice question answering across 22 languages. The multilingual calibration studies
cited in the main text repair the probabilities with temperature scaling on target-language data or small translated
calibration sets \citep{arxiv2210_12265,arxiv2311_08669_ax}, and \citet{feng2024abstain_tc} propose multilingual
feedback, which reduces the abstention gap. \citet{xin2021abstention} study selective prediction for NLP models, and
\citet{kamath2020selective} showed that abstention based on softmax probabilities fares poorly under domain shift.
Conformal risk control and Learn-then-Test \citep{angelopoulos2024crc,angelopoulos2025ltt} give finite-sample
guarantees for threshold selection that our empirical fit lacks, and group-conditional risk control
\citep{arxiv2607_24562_ax} would give such a guarantee per group. \citet{arxiv2605_14260_ax} also show that pooled
conformal calibration can distort group-wise coverage. In concurrent work on security decisions with this interface,
\citet{arxiv2609_33401_ax} reports that a frozen policy validated under an error budget exceeds it on unseen inputs.
Its groups are sets of related security-benchmark inputs, where ours are languages. Two further studies of the class
report the same held-out miss \citep{arxiv2610_02267_ax,arxiv2609_33843_ax}. \citet{arxiv2610_19811_ax} find
that naive thresholds on models of the class exceed their risk target in about half of settings, and Learn-then-Test
thresholds in 7.8\%. \citet{arxiv2610_11392_ax} report that a hosted model of the class is sensitive to wording and
systematically underconfident. \citet{arxiv2609_36965_ax}, also
concurrent, address the language difference on the training side with a Chinese-language model of the class, whereas
we measure what a fixed gate does with it.
\end{document}